# Brain age identification from diffusion MRI synergistically predicts neurodegenerative disease

Chenyu Gao,[1] Michael E. Kim,[2] Karthik Ramadass,[1] Praitayini Kanakaraj,[2] Aravind R. Krishnan,[1] Adam M. Saunders,[1] Nancy R. Newlin,[2] Ho Hin Lee,[2] Qi Yang,[2] Warren D. Taylor,[3,4] Brian D. Boyd,[3] Lori L. Beason-Held,[5] Susan M. Resnick,[5] Lisa L. Barnes,[6] David A. Bennett,[6] Marilyn S. Albert,[7] Katherine D. Van Schaik,[8] Derek B. Archer,[9,10] Timothy J. Hohman,[9,10] Angela L. Jefferson,[9,10,11] Ivana Išgum,[12] Daniel Moyer,[2] Yuankai Huo,[1,2] Kurt G. Schilling,[8] Lianrui Zuo,[1] Shunxing Bao,[1] Nazirah Mohd Khairi,[1] Zhiyuan Li,[1] Christos Davatzikos,[13] Bennett A. Landman[1,2,3,8,10]

# Abstract

Estimated brain age from magnetic resonance image (MRI) and its deviation from chronological age can provide early insights into potential neurodegenerative diseases, supporting early detection and implementation of prevention strategies to slow disease progression and onset. Diffusion MRI (dMRI), a widely used modality for brain age estimation, presents an opportunity to build an earlier biomarker for neurodegenerative disease prediction because it captures subtle microstructural changes that precede more perceptible macrostructural changes. However, the coexistence of macro- and micro-structural information in dMRI raises the question of whether current dMRI-based brain age estimation models are leveraging the intended microstructural information or if they inadvertently rely on the macrostructural information. To develop a microstructure-specific brain age, we propose a method for brain age identification from dMRI that mitigates the model's use of macrostructural information by non-rigidly registering all images to a standard template. Imaging data from 13,398 participants across 12 datasets were used for the training and evaluation. We compare our brain age models, trained with and without macrostructural information mitigated, with an architecturally similar T1-weighted (T1w) MRI-based brain age model and two recent, popular, openly available T1w MRI-based brain age models that primarily use macrostructural information. We observe difference between our dMRI-based brain age and T1w MRI-based brain age across stages of neurodegeneration, with dMRI-based brain age being older than T1w MRI-based brain age in participants transitioning from cognitively normal (CN) to mild cognitive impairment (MCI) (p-value = 0.023), but younger in participants already diagnosed with Alzheimer's disease (AD) (p-value < 0.001). Classifiers using T1w MRI-based brain ages generally outperform those using dMRI-based brain age in classifying CN vs. AD participants. Conversely, dMRI-based brain age may offer advantages over T1w MRI-based brain age in predicting the transition from CN to MCI.

**Author affiliations:**
[1] Department of Electrical and Computer Engineering, Vanderbilt University, Nashville, Tennessee 37240, USA
[2] Department of Computer Science, Vanderbilt University, Nashville, Tennessee 37240, USA
[3] Vanderbilt Center for Cognitive Medicine, Department of Psychiatry and Behavioral Sciences, Vanderbilt University Medical Center, Nashville, TN, USA
[4] Geriatric Research, Education, and Clinical Center, Veterans Affairs Tennessee Valley Health System, Nashville, TN, USA
[5] Laboratory of Behavioral Neuroscience, National Institute on Aging, National Institutes of Health, Baltimore, Maryland, USA
[6] Rush Alzheimer's Disease Center, Rush University Medical Center, Chicago, IL
[7] Department of Neurology, Johns Hopkins University School of Medicine, Baltimore, MD, USA
[8] Department of Radiology and Radiological Sciences, Vanderbilt University Medical Center, Nashville, USA
[9] Vanderbilt Memory and Alzheimer's Center, Vanderbilt University Medical Center, Nashville, Tennessee 37240, USA
[10] Department of Neurology, Vanderbilt University Medical Center, Nashville, Tennessee 37240, USA
[11] Department of Medicine, Vanderbilt University Medical Center, Nashville, Tennessee 37240, USA
[12] Department of Biomedical Engineering and Physics, Department of Radiology and Nuclear Medicine, Amsterdam University Medical Center, University of Amsterdam, Amsterdam, Netherlands
[13] AI2D Center for AI and Data Science, University of Pennsylvania, Philadelphia, Pennsylvania 19104, USA

Correspondence to: Chenyu Gao
Full address: 2301 Vanderbilt Pl., PO Box 351679 Station B, Nashville, TN 37235-1679
E-mail: chenyu.gao@vanderbilt.edu

# Introduction

Patterns of macro- and micro-structural changes associated with normal brain aging can be captured from magnetic resonance images (MRIs) by machine learning methods to construct brain age—an important imaging biomarker in the fields of neuroscience and radiology. (Wen et al., 2024) By comparing an individual's MRI-derived brain age with their chronological age, deviations from the normal aging trajectory can be identified. A brain age that is less advanced than the chronological age may reflect good brain health and resilient aging. (Cole et al., 2019; Kolbeinsson et al., 2020; Liew et al., 2023) Conversely, a brain age that is more advanced than the chronological age may reflect accelerated aging, which could be indicative of neurodegenerative diseases. (Bashyam et al., 2020; Davatzikos et al., 2009; Liem et al., 2017) Early identification of at-risk individuals enables proactive management of conditions like mild cognitive impairment (MCI) or Alzheimer's disease (AD), leading to timely and targeted therapeutic strategies, which may slow disease progression. (Bateman et al., 2012; Livingston et al., 2020)

Specificity and sensitivity are two critical aspects of machine learning models in clinical applications. In the context of brain age estimation—where the difference between estimated brain age and chronological age (i.e., the brain age gap) can be used to classify whether an individual is developing neurodegenerative diseases—specificity relates to accurate chronological age estimation for individuals who are neither experiencing nor on a trajectory to develop neurodegenerative diseases or cognitive decline, to avoid false alarms. Sensitivity, on the other hand, relates to detecting deviations from the normal aging trajectory, as indicated by large positive brain age gaps in individuals who are either experiencing or on a trajectory to develop neurodegenerative disease or cognitive decline. Ideally, we would hope to detect such deviations well before clinical diagnosis, allowing ample time for intervention.

Considerable efforts have been made to enhance the specificity of brain age estimation. Among these efforts, four trends stand out. First, there is a growing emphasis on using large datasets that encompass a diverse range of cohorts, characterized by variations in age, race/ethnicity, sex, education, and geographic location, as well as acquisitions that differ in scanner type, imaging parameters, and quality. (Bashyam et al., 2020; Dufumier et al., 2022; Wood et al., 2022) The rationale for using larger and more heterogeneous datasets is to develop models that are robust and generalizable, capable of maintaining accuracy when applied to previously unseen data. Second, the field is witnessing a paradigm shift towards the adoption of deep neural networks with sophisticated architectural designs. (Bashyam et al., 2020; H. J. Cai et al., 2023; Cheng et al., 2021; Dinsdale et al., 2021; Gao et al., 2024; Wood et al., 2022) These networks have the capacity to learn complex feature representations directly from brain images, offering an advantage over traditional machine learning models that rely on hand-crafted and preselected features. (Davatzikos et al., 2009; Franke et al., 2010; Habes et al., 2016; Wen et al., 2024) Third, the fusion of multimodal imaging data is increasingly being used. (H. J. Cai et al., 2023; Liem et al., 2017) By combining data from different imaging modalities, models can potentially capture a wider spectrum of age-related changes. Fourth, transfer learning is being used to leverage pre-trained models on large datasets to improve performance on smaller, target datasets. (C.-L. Chen et al., 2020; Wood et al., 2024) Through these efforts, the field has reported progressively lower mean absolute errors in brain age estimation for healthy individuals.

Comparatively, fewer efforts have been directed towards improving the sensitivity of brain age estimation. (Bashyam et al., 2020; Davatzikos et al., 2009; Habes et al., 2016; Wen et al., 2024) A common theme of brain age estimation methods involves the use of T1-weighted (T1w) images, which primarily capture macrostructural and intensity information. (Bashyam et al., 2020; Cheng et al., 2021; Dinsdale et al., 2021) T1w images allow us to observe changes related to brain aging, such as atrophy, (Fotenos et al., 2005; Ridha et al., 2006) cortical thinning, (Bakkour et al., 2009) ventricular enlargement, (Ferrarini et al., 2006; Nestor et al., 2008) and white matter hyperintensities. (De Groot et al., 2002; Habes et al., 2016) However, T1w images lack detailed information about white matter regions, making them less sensitive to the early microstructural changes that precede noticeable macrostructural changes. (DeIpolyi et al., 2005; Kantarci et al., 2005; Müller et al., 2005; Ringman et al., 2007; Weston et al., 2015) With regard to MCI and AD, emerging evidence highlights distinct white matter abnormalities, including axonal loss, (Sachdev et al., 2013) demyelination, (Ihara et al., 2010) and microglial activation. (Simpson et al., 2007) Importantly, these changes manifest up to 22 years prior to symptom onset (Lee et al., 2016; Nasrabady et al., 2018) and have independent contributions to cognitive decline beyond that of hippocampal volume. (Archer et al., 2020) Diffusion MRI (dMRI), on the other hand, can capture white matter microstructural alterations, offering the potential to develop an earlier biomarker for neurodegenerative disease prediction. (S. R. Cox et al., 2016; Kantarci et al., 2005; Müller et al., 2005; Ringman et al., 2007) Nonetheless, the presence of macrostructural information within dMRI data presents a confounding factor. It remains unclear whether current brain age estimation models based on dMRI data are leveraging the intended microstructural information or if they are inadvertently relying on macrostructural information.

In this study, we isolate the microstructural information from dMRI data for brain age estimation. Specifically, we use nonrigid (deformable) registrations to warp all brains to one standard template brain, thereby mitigating macrostructural variations across the dataset. We hypothesize that the microstructure-informed brain age will serve as an earlier biomarker for neurodegenerative diseases, offering improved predictive capabilities for conditions such as MCI. To serve the testing of the hypothesis, we included 12 datasets comprising a total of 13,398 participants, with longitudinal data included. For the architecture of our brain age estimation models, we used 3D residual neural network (ResNet), (K. He et al., 2016) a well-established convolutional neural network architecture in the field. To compare microstructure-informed brain age with "micro- and macro-structure mixture"-informed brain age, we trained the ResNets using dMRI-derived data with and without the macrostructural information mitigated through non-rigid registrations. Additionally, to compare microstructure-informed brain age with "T1w macrostructure"-informed brain age, we also trained two separate T1w-based brain age estimation models. One model uses the same ResNet architecture, while the other uses an open-source architecture known as TSAN, which was reported to achieve low estimation error (Cheng et al., 2021); both were trained on the same set of participants as the dMRI-based models. For a more comprehensive comparison, we also applied DeepBrainNet (Bashyam et al., 2020), another highly regarded T1w-based brain age estimation model, to our data using pretrained model weights. We conducted comparisons of these brain ages (Fig. 1). We examined their differences across diagnostic groups, such as cognitively normal (CN), AD, MCI, and CN participants who later transitioned to MCI. We assessed their performance in classifying participants within these groups and in predicting whether a CN participant will transition to MCI in the future, from 0 to 9 years prior to diagnosis. Furthermore, we investigated the added value of microstructure-informed brain age on T1w-based brain ages in predicting MCI incidence in survival analysis.

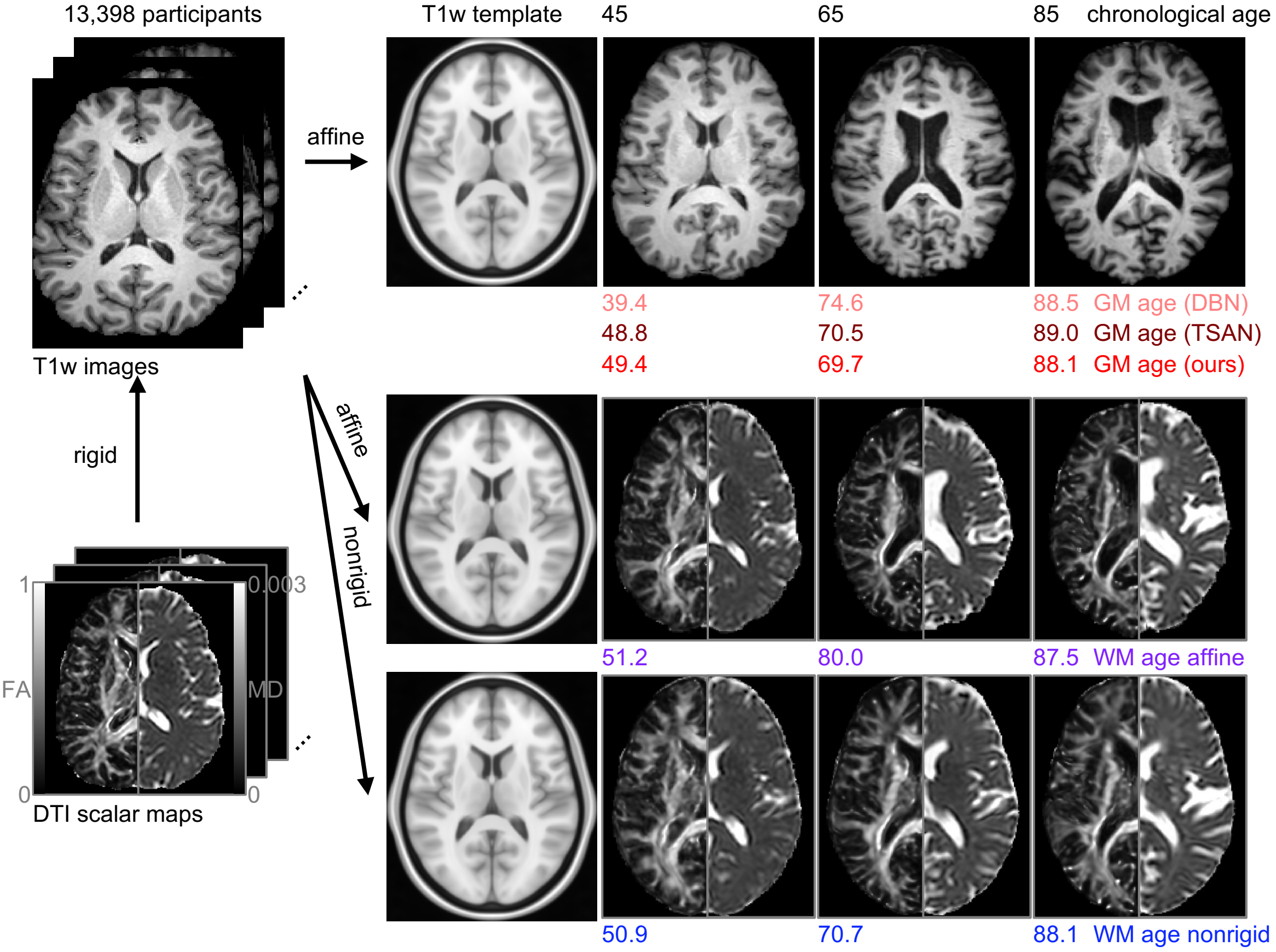

**Figure 1.** Brain age estimation frameworks have proven effective for using affinely aligned brain images to identify common patterns of aging, with deviations from these patterns likely indicating presence of abnormal neuropathologic processes. A common theme of existing brain age estimation methods has been using T1w MRI, denoted as "GM age" in the first row. Among them, there have been many innovations in network design, such as DeepBrainNet (DBN) (Bashyam et al., 2020) and the 3D convolutional neural network of TSAN (Cheng et al., 2021). T1w MRI lacks detail in white matter (WM). Here, we take the two most commonly used modalities for characterizing WM microstructure, fractional anisotropy (FA), and mean diffusivity (MD), and we evaluate brain age estimation in two contexts. First, we examine the direct substitution of FA and MD for T1w image, which we denote as "WM age affine" in the second row. A substantial amount of macrostructural differences is still present in WM age affine, notably ventricle enlargement. To isolate the microstructural changes, we apply non-rigid (deformable) registration into template space to mitigate the macrostructural changes and produce the "WM age nonrigid" in the third row. We explore the relative timing of changes in these brain age variants and their relative explainability in the context of mild cognitive impairment. Throughout the paper, we adhere to a consistent color scheme when visualizing results from different brain age estimates within the same plot to facilitate easier visual inspection. Specifically, we use red to represent GM ages, blue for WM age nonrigid, and purple for WM age affine.

# Materials and methods

## Datasets

We included 12 datasets. After quality assurance, there were a total of 13,398 participants, contributing to 18,673 imaging sessions that included both cross-sectional and longitudinal data. For every imaging session, both diffusion MRI and T1w MRI were acquired. We selected 10,647 CN participants and divided them into five folds for training and cross-validation. During training, data samples from longitudinal sessions and multiple scans were included and treated as a form of data augmentation. To avoid biasing the model toward participants who have more data samples, we normalized each sample's probability of being sampled at each iteration by the total number of samples belonging to that participant. For example, consider a training set with only two participants: A (who has two samples, $d_{A1}$ and $d_{A2}$), and B (who has one sample, $d_{B1}$). After normalization, the probabilities of sampling $d_{A1}$, $d_{A2}$, and $d_{B1}$ are 0.25, 0.25, and 0.5, respectively. The remaining 2,751 participants were held out as the testing set. IRB of Vanderbilt University waived ethical approval for de-identified access of the human subject data.

**Table 1. Dataset characteristics.**

| | n participants (male/female) | n participants | n sessions | follow-up interval[a] | n scans[b] | age mean ± std | age range |
|---|---|---|---|---|---|---|---|
| ADNI (Jack Jr. et al., 2008) | 1250 (599/651) | 591 CN, 440 MCI, 219 AD | 3015 (1370 CN, 1204 MCI, 441 AD) | 1.3 ± 0.9 | 3177 | 74.6 ± 7.5 | 50.5 - 95.9 |
| BIOCARD (Sacktor et al., 2017) | 186 (70/116) | 142 CN, 39 MCI, 5 AD | 404 (336 CN, 63 MCI, 5 AD) | 2.1 ± 0.4 | 805 | 71.3 ± 8.2 | 34.0 - 93.0 |
| BLSA (Shock, 1984) | 1026 (471/555) | 964 CN, 39 MCI, 23 AD | 2698 (2579 CN, 80 MCI, 39 AD) | 2.2 ± 1.2 | 5312 | 73.3 ± 13.5 | 22.4 - 103.0 |
| HCPA (Bookheimer et al., 2019; Harms et al., 2018) | 719 (316/403) | 719 CN | 719 (719 CN) | N/A | 719 | 60.4 ± 15.7 | 36.0 - 100.0 |
| ICBM (Kötter et al., 2001) | 184 (83/101) | 184 CN | 184 (184 CN) | N/A | 218 | 41.5 ± 15.6 | 19.0 - 80.0 |
| NACC (Beekly et al., 2007) | 638 (254/384) | 459 CN, 25 MCI, 154 AD | 673 (491 CN, 26 MCI, 156 AD) | 2.0 ± 1.2 | 728 | 68.4 ± 11.3 | 43.5 - 100.1 |
| OASIS3 (LaMontagne et al., 2019) | 249 (119/130) | 204 CN, 19 MCI, 26 AD | 259 (214 CN, 19 MCI, 26 AD) | 1.7 ± 0.7 | 364 | 72.3 ± 7.8 | 46.2 - 92.2 |
| OASIS4 (Koenig et al., 2020) | 90 (50/40) | 13 CN, 10 MCI, 67 AD | 90 (13 CN, 10 MCI, 67 AD) | N/A | 91 | 76.0 ± 9.1 | 50.8 - 94.1 |
| ROSMAPMARS (Bennett et al., 2018; L Barnes et al., 2012) | 642 (124/518) | 474 CN, 148 MCI, 20 AD | 1342 (1137 CN, 184 MCI, 21 AD) | 2.4 ± 0.8 | 1342 | 81.2 ± 7.3 | 58.8 - 102.9 |
| UK BioBank (Sudlow et al., 2015) | 7777 (3630/4147) | 7777 CN | 7777 (7777 CN) | N/A | 7777 | 64.3 ± 7.6 | 46.1 - 82.8 |
| VMAP (Jefferson et al., 2016) | 296 (171/125) | 168 CN, 128 MCI | 857 (507 CN, 350 MCI) | 1.8 ± 0.6 | 857 | 74.8 ± 7.2 | 60.4 - 96.0 |
| WRAP (Johnson et al., 2018) | 341 (111/230) | 335 CN, 4 MCI, 2 AD | 555 (549 CN, 4 MCI, 2 AD) | 2.5 ± 1.2 | 555 | 62.5 ± 6.7 | 44.3 - 76.7 |
| Combined | 13398 (5998/7400) | 12030 CN, 852 MCI, 516 AD | 18573 (15876 CN, 1940 MCI, 757 AD) | 1.9 ± 1.1 | 21945 | 69.6 ± 11.7 | 19.0 - 103.0 |

CN = cognitively normal; MCI = mild cognitive impairment; AD = Alzheimer's disease. The numbers reflect the datasets following quality assurance and do not correspond to the characteristics of the original datasets. [a]The time between consecutive longitudinal sessions. The unit for follow-up interval and age is year. [b]The number of diffusion MRI scans is reported. The number of T1w MRI scans varies.

## Data preprocessing

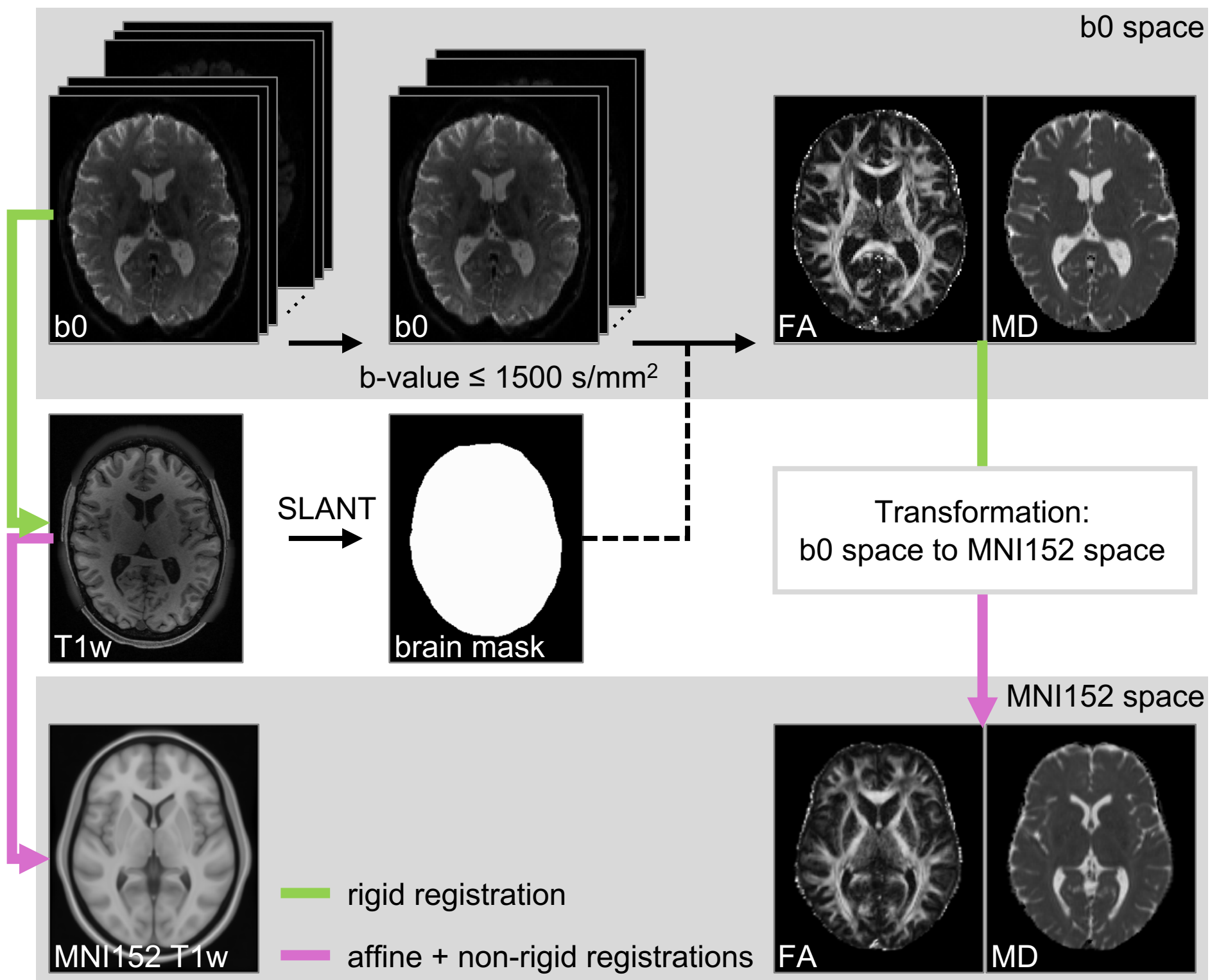

**Figure 2.** The fractional anisotropy (FA) and mean diffusivity (MD) images are calculated from volumes with b-value ≤ 1500 s/mm$^2$ extracted from preprocessed diffusion MRI data. Rigid registration (the green line) between b0 image and T1w image, and affine and non-rigid (deformable) registrations (the purple line) between T1w image and MNI152 T1w template are performed and concatenated to form the transformation from b0 space to MNI152 space. A brain mask is computed from T1w image with SLANT whole brain segmentation pipeline and applied to the FA and MD images.

For dMRI data, we used PreQual, (L. Y. Cai et al., 2021) an end-to-end preprocessing pipeline, for denoising and to attenuate susceptibility artifacts, motion, and eddy current artifacts. We computed two diffusion tensor imaging (DTI) scalar maps, fractional anisotropy (FA) and mean diffusivity (MD), from the volumes acquired with a b-value ≤ 1500 s/mm$^2$ and transformed them to MNI152 space (Fonov et al., 2011) (Fig. 2). There are two types of transformations: the first involves a rigid transformation (from b0 to T1w) followed by an affine transformation (from T1w to MNI152), which aligns the brain to the template while preserving macrostructural variations. The second type combines the rigid and affine transformations with a nonrigid (deformable) transformation, (Avants et al., 2008) further warping the brain to match the template and mitigate macrostructural variations (Fig. S1). For T1w images, we applied only the affine transformation to preserve macrostructural information. The registrations were performed using ANTs. (Avants et al., 2011) The nonrigid registration was done with the deformable SyN option. (Avants et al., 2008) We used SLANT-TICV, (Huo et al., 2019; Liu et al., 2022) a deep learning-based whole brain segmentation pipeline, to generate brain masks for skull-stripping the FA, MD, and T1w images.

The registered and skull-stripped images were then downsampled and cropped to 128×152×128 with an isotropic resolution of 1.5 mm$^3$ to reduce GPU RAM requirements. Since TSAN (Cheng et al., 2021) was originally trained on a relatively smaller dataset, (Cheng et al., 2021) we retrained it on our richer dataset to enable a fair comparison. We followed the requirements described in the paper (Cheng et al., 2021) for retraining TSAN. The images were downsampled and cropped to 91×109×91 with an isotropic resolution of 2 mm$^3$. For running the pretrained DeepBrainNet model, we strictly followed the preprocessing steps and software tools described in the paper. (Bashyam et al., 2020) The preprocessed images were manually checked. Table 1 reflects what remained after the quality assurance.

## Brain age estimation models

We included three types of models, each type represented as a row in Fig. 1. The first type represents T1w MRI-based models. Since the images capture high-contrast structural information about gray matter (GM) regions, we name these models "GM age" models. Among them, we have our model ("GM age (ours)"), which uses a 3D ResNet (K. He et al., 2016) as the architecture and takes the T1w image, along with sex and race information, as input. The embedding from the convolutional layers is concatenated with the vectorized sex and race information (one-hot encoded) before entering the fully connected layers to output the estimated brain age. We also included TSAN ("GM age (TSAN)"), as a comparison with an architecture that achieved low estimation error. (Cheng et al., 2021) TSAN uses a two-stage cascade network architecture, where the first-stage network estimates a rough brain age, and the second-stage network refines the brain age estimate. (Cheng et al., 2021) TSAN takes the T1w image and sex information as input. Both GM age (ours) and GM age (TSAN) were trained from scratch on our skull-stripped T1w images affinely registered to the MNI152 template. Additionally, we included the pretrained DeepBrainNet (Bashyam et al., 2020) ("GM age (DBN)") as another T1w MRI-based method. DeepBrainNet uses a 2D convolutional neural network and was pretrained on a large dataset (N=11,729). (Bashyam et al., 2020) It uses only the T1w image as input. The inference process for GM age (DBN) strictly followed the processing steps described in the paper. (Bashyam et al., 2020) The second type uses a similar architecture to the 3D ResNet of GM age (ours), except that it substitutes the T1w image with FA and MD images, skull-stripped and affinely registered to the MNI152 template. The FA and MD images are concatenated together as two channels before being fed into the network. Because the input images capture microstructural information, which has the most variation in white matter (WM) regions, we name the model "WM age affine". The third type, "WM age nonrigid", uses the exact same model architecture as "WM age affine", except that the input images are skull-stripped FA and MD images non-rigidly registered to the MNI152 template.

We trained the models, except "GM age (DBN)", using scans of individuals aged between 45 and 90 years. This range provided a sufficiently large sample of midlife to older adults, aligning with our goal of investigating age-related changes linked to MCI and AD. We excluded scans of individuals outside this range because their numbers were relatively small for both training and evaluation. We implemented two strategies to mitigate the models' bias towards middle-aged participants. First, the age of the scan is sampled uniformly during training. Scans being sampled are assigned decayed probabilities of being sampled again, ensuring all available scans can be iterated through in fewer iterations. Second, we fit bias correction parameters (slope and intercept) on the validation set (one of the five folds of the training set) after model training and apply the correction to the estimated brain ages following the steps described in detail in the paper. (Smith et al., 2019) For "GM age (DBN)", the bias correction parameters are computed from the entire training set.

## Classification of MCI/AD participants

To determine whether the estimated brain age by each model is indicative of neurodegeneration, we perform classification of participants by cognitive status. The features used for classification include sex, chronological age, and brain age gap (BAG), which we define as the difference between the estimated brain age and the chronological age (Eq. 1). For participants with longitudinal sessions, we compute the change rate of the brain age gap by taking the difference between the brain age gaps from two adjacent sessions and dividing it by the interval. We generate additional features by computing interactions with chronological age and sex.

$$BAG = Age_{estimated} - Age_{chronological} \tag{1}$$

We separate participants into four groups for classification. The first group consists of CN participants who remain CN in the follow-up sessions. The second and third groups include participants who are diagnosed with MCI and AD, respectively. The fourth group comprises participants who are CN in the current session but will transition to MCI in future sessions, which we define as "CN*". We apply a greedy algorithm to obtain matched and balanced data points for group comparison and classification. Specifically, when comparing multiple ($N \geq 2$) groups of participants, we iteratively search for data points of unused participants, one from each group, that have the same sex and the closest age, with the age difference not exceeding one year. Additionally, when matching CN and CN* data, the time to the last CN session (for the CN data point) and the time to the first MCI diagnosis (for the CN* data point) must also match, with a difference of no more than one year. The resulting groups contain only one data point for each matched participant. We use three different machine learning classifiers for the classification: logistic regression, linear support vector machine (SVM), and random forest. The input features are min-max normalized to the range of -1 to 1. Missing values for each feature are imputed with the mean value of that feature.

## Prediction of transition from CN to MCI

To understand the translational impact of our WM age nonrigid model, we conduct a prediction experiment to determine whether brain age can predict the future transition of a CN participant to MCI. We use sliding windows (with window length of one year and stride of 0.5 year) to sample data points at various time points (T-0, T-1, …, T-n) before the first MCI diagnosis and assess the classifiers' ability to differentiate these data points from matched CN data points using brain age-derived features. The prediction experiment is structured into two setups, each with a distinct experimental procedure and underlying logic.

In the first setup, which is called the "global model" approach, we use the greedy algorithm to match CN data points with those transitioning to MCI (CN*). We then apply leave-one-out cross-validation, where we train classifiers on the remaining data and test them on the left-out participant and their matched CN data points. This process is repeated for all CN* participants. Subsequently, we slide the window across the "time to MCI" axis, select the most central data point pair from each participant, and use bootstrapping to compute the mean and 95% confidence intervals of the area under the receiver operating characteristic curve (AUC) within the window.

In the second setup, which is called the "time-specific models" approach, we slide the window across the "time to MCI" axis to create subsets of data, each representing a different "time to MCI" range. For each subset, we match CN data points using the greedy algorithm, perform leave-one-out cross-validation, and record the predicted probabilities. We then bootstrap to compute the mean and 95% confidence intervals of the AUC for each subset. This approach utilizes multiple models, each tailored to a specific "time to MCI" range.

Flowcharts illustrating both setups are provided in the Supplementary Materials. We include both setups because they provide complementary perspectives on the predictive utility of the brain age models. The global model approach provides an overall assessment across the entire pre-diagnosis period, while the time-specific models reveal how predictive accuracy changes as participants near MCI onset. Together, the two setups offer insight into the temporal dynamics of our predictive signal.

## Survival analysis

To assess whether WM age nonrigid provides additional predictive value over GM ages for the incidence of MCI, we conduct survival analysis. Our cohort for this analysis includes baseline sessions from 131 CN* participants. We also incorporate baseline sessions from 290 participants within the same datasets who remained CN until their last recorded session. The diagnosis of MCI is treated as the event of interest, with all other observations considered censored. We use Cox proportional-hazards models (D. R. Cox, 1972) to evaluate the risk factors associated with MCI onset. Our analysis is structured into two scenarios: the first excluded WM age nonrigid, fitting models with chronological age, sex, and GM ages as covariates, while the second included WM age nonrigid alongside the covariates used in the first scenario.

## Statistical analysis

For testing the null hypothesis that two related paired samples come from the same distribution, we use the Wilcoxon signed-rank test. Accuracy and AUC are reported for classification and prediction performance. The concordance index (C-index) is reported for the Cox proportional-hazards models. Bootstrapping ($n$=1000) is used to calculate the mean and 95% confidence intervals of these metrics. To assess fit of the Cox proportional-hazards models, we report the Akaike information criterion (AIC) scores. We evaluate improvements in model fit using the likelihood ratio test, which compares the log-likelihoods of the nested models with and without the inclusion of WM age nonrigid. The chi-squared ($\chi^2$) statistic and corresponding p-value are computed to determine the statistical significance of the improvements with the addition of WM age nonrigid. We choose an a priori threshold of p-value < 0.05 to denote statistical significance.

# Results

## Brain age estimation of five models

The $\overline{|BAG|}$ (mean of absolute BAG) is greater in the AD group than in the MCI group, and greater in the MCI group than in the CN group (Fig. 3). This trend is also reflected in the density plots of BAG versus chronological age: in the AD group, the density distribution appears narrower and more diagonally sloped in an elliptical shape compared to MCI, and in the MCI group, it appears narrower and more diagonally sloped in an elliptical shape compared to CN. In the CN* group, the $\overline{|BAG|}$ for all models–with the exception of GM age (DBN)–showed an increase when compared to the CN group. For example, the $\overline{|BAG|}$ of WM age nonrigid rose from 3.21 years in the CN group to 3.52 years in the CN* group. Among the CN participants, GM age (ours) and GM age (TSAN) achieved the lowest $\overline{|BAG|}$ (~3.1 years).

## Difference between WM age and GM age across stages of neurodegeneration

In our matched dataset, controlled for age, sex, and time-to-event, we found significant differences between WM age nonrigid and GM age (ours) among CN* participants (Fig. 4). In this group, WM age nonrigid exceeded GM age (ours) by an average of 0.48 years. A more pronounced difference was observed in participants with AD, where WM age nonrigid was, on average, 0.99 years lower than GM age (ours) (p-value < 0.001). No significant differences were detected between WM age nonrigid and GM age (ours) in those who remained CN across all available sessions or those who were classified as MCI. The p-values were obtained using the Wilcoxon signed-rank test.

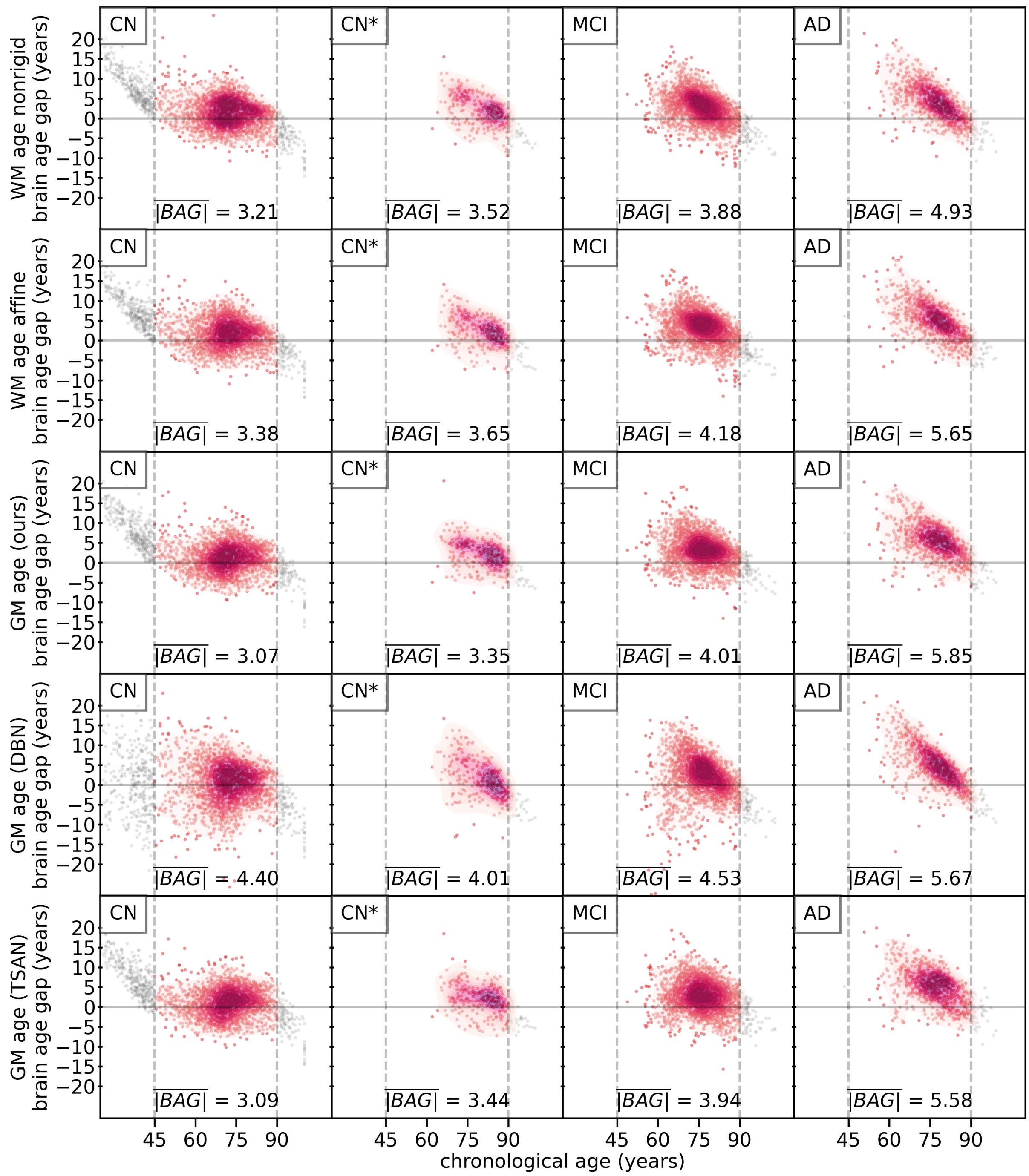

**Figure 3.** As neurodegeneration progresses, estimated brain age generally deviates more from chronological age, as reflected by the shape of the density distribution of the brain age gap (BAG, which equals estimated brain age minus chronological age) and the $\overline{|BAG|}$ value. CN* are participants cognitively normal at present but diagnosed with MCI in follow-up sessions. Scatters outside the training age range (45 to 90 years) are colored gray and shown only to illustrate the poor estimation performance on out-of-distribution data. These out-of-distribution data points were excluded from calculation of $\overline{|BAG|}$ and from subsequent analyses.

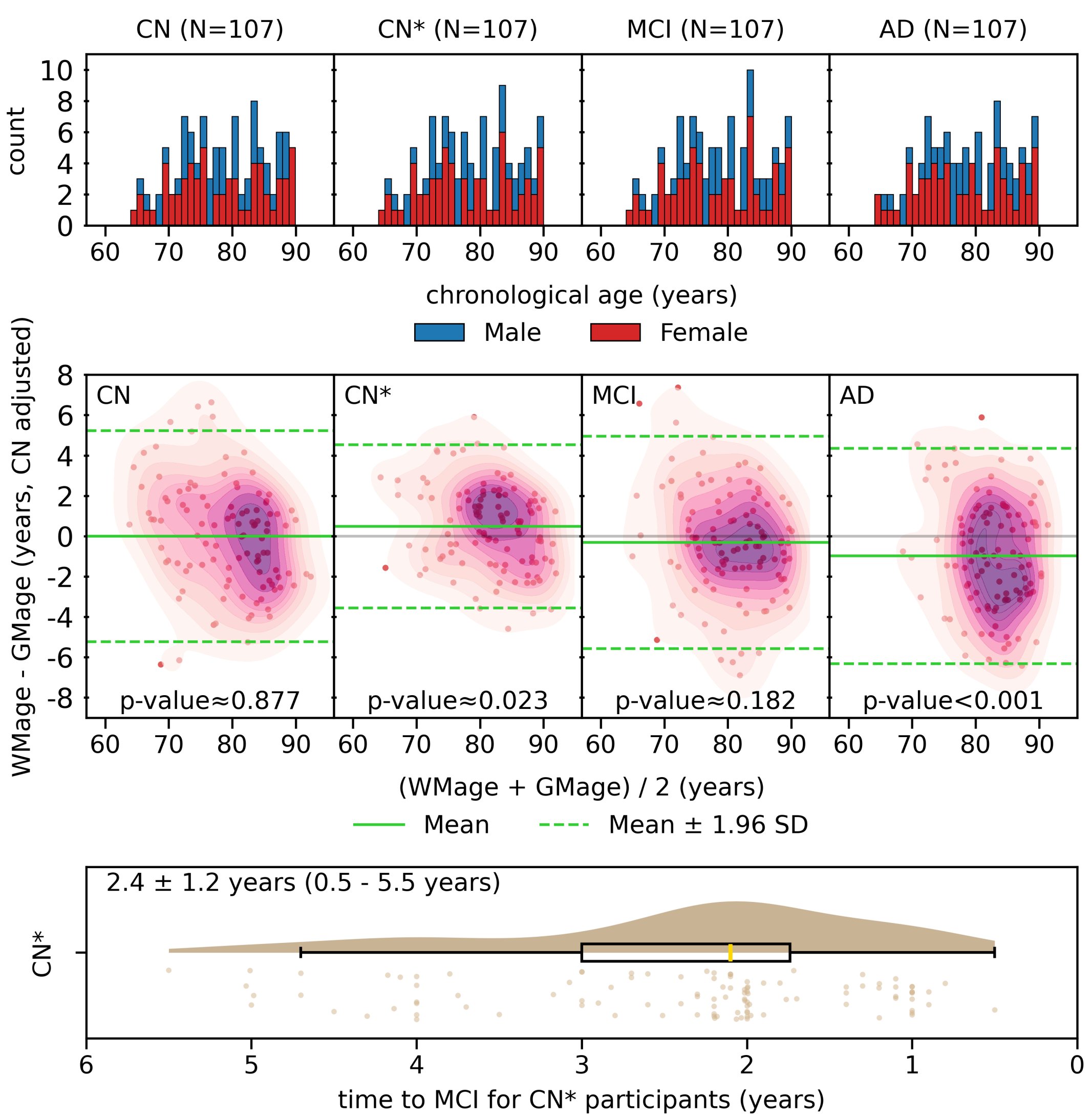

**Figure 4.** Data points from four diagnosis groups are matched regarding age and sex (and time to last CN and time to first MCI for matching CN and CN* data points). The differences between WM age nonrigid and GM age (ours) are adjusted by the mean of the differences for the CN group. Wilcoxon signed-rank tests show significant difference between WM age nonrigid and GM age (ours) on both CN* and AD participants.

**Table 2. Classification of CN vs. AD, CN vs. MCI, and CN vs. CN* using chronological age, sex, and brain age-related features.** To facilitate easier visual inspection, we use blue for WM age nonrigid, purple for WM age affine, red for GM age (ours), and green for combinations of GM age (ours) and WM age nonrigid. The same color scheme is followed in other figures. The highest AUCs across the three feature sets (excluding "basic" and "basic + GM age (ours) + WM age nonrigid") are highlighted in bold. Results based on Linear SVM are presented here while the full version can be found in the Supplementary Materials.

| | CN vs. AD (N=458 matched pairs) | | CN vs. MCI (N=694 matched pairs) | | CN vs. CN* (N=118 matched pairs) | |
|---|---|---|---|---|---|---|
| Features | Accuracy | AUC | Accuracy | AUC | Accuracy | AUC |
| basic: chronological age + sex | 0.50 (0.50, 0.50) | 0.50 (0.50, 0.50) | 0.50 (0.50, 0.50) | 0.50 (0.50, 0.50) | 0.50 (0.50, 0.50) | 0.50 (0.50, 0.50) |
| basic + WM age nonrigid | 0.65 (0.63, 0.67) | 0.72 (0.70, 0.74) | 0.60 (0.59, 0.62) | **0.65 (0.62, 0.67)** | 0.62 (0.57, 0.67) | **0.64 (0.59, 0.69)** |
| basic + WM age affine | 0.67 (0.65, 0.70) | 0.74 (0.72, 0.76) | 0.60 (0.58, 0.62) | 0.63 (0.61, 0.65) | 0.59 (0.54, 0.64) | 0.63 (0.58, 0.69) |
| basic + GM age (ours) | 0.69 (0.66, 0.71) | **0.76 (0.73, 0.78)** | 0.60 (0.58, 0.62) | 0.64 (0.61, 0.66) | 0.62 (0.57, 0.67) | 0.63 (0.57, 0.69) |
| basic + GM age (ours) + WM age nonrigid | 0.70 (0.67, 0.72) | 0.75 (0.73, 0.78) | 0.63 (0.61, 0.65) | 0.66 (0.64, 0.68) | 0.61 (0.56, 0.65) | 0.64 (0.59, 0.69) |

CN = cognitively normal; AD = Alzheimer's disease; MCI = mild cognitive impairment; CN* = cognitively normal at present but diagnosed with mild cognitive impairment in the future.

## Classification of cognitively normal vs. current and future mild cognitive impairment/Alzheimer's disease participants

We conducted three classification tasks to differentiate between CN participants and those with AD, MCI, and CN participants who would later transition to MCI (CN*) (Table 2 and Supplementary Materials). Linear classifiers (logistic regression and linear SVM) show baseline accuracy and AUC of 0.5 with chronological age and sex, confirming that the samples are matched for these variables. As the classification task shifted from distinguishing CN vs. AD to CN vs. MCI, we observed an increase in the difficulty of classification, as reflected by decreased accuracy and AUC. In the CN vs. AD task, features derived from GM ages generally outperform those from WM age nonrigid. However, in the CN vs. MCI task, the performance gap between GM age and WM age nonrigid features narrowed. In the task of classifying CN vs. CN* participants, features derived from WM age nonrigid marginally outperform those from GM ages, although the difference is not statistically significant. Notably, combining WM age nonrigid features with GM age features consistently results in the best performance across all classification tasks.

## Prediction of transition from cognitively normal to mild cognitive impairment from 1, …, *n* years pre-diagnosis

Data points of 131 participants, who had imaging data acquired from periods when they were CN and subsequent periods when they transitioned to MCI, were matched with those from CN participants (Fig. S2). At the time of MCI diagnosis (T-0), all feature combinations exhibited similar performance levels (Fig. S3). Features derived from WM ages exhibited a slight advantage across all three types of classifiers—logistic regression, linear SVM, and random forest—and both the global model and time-specific models, although the differences were not statistically significant. From T-0 to T-4 (0 to 4 years before MCI diagnosis), features derived from WM age nonrigid and WM age affine, as well as their combinations with other brain age-derived features, show advantages over other features. Specifically, under the global model setup, random forest classifiers showed that WM age affine-derived features yielded the highest performance in the first half of this four-year period (0 to 2 years before MCI diagnosis), with an AUC of 0.7. In contrast, during the latter half (2-4 years before MCI diagnosis), WM age nonrigid-derived features achieved the best performance, with an AUC of 0.76.

At T-5 (5 years before MCI diagnosis), features derived from GM age (DBN) outperformed other features when using random forest classifiers under the global model setup, achieving an AUC of 0.78; logistic regression and linear SVM classifiers using features derived from GM ages showed comparable performance to those using features derived from WM age affine.

## Added value of WM age nonrigid in predicting mild cognitive impairment incidence

Among the 421 participants included for the survival analysis, 131 progressed to MCI, while the remaining 290 remained cognitively normal. The detailed survival table can be found in the supplementary materials (Table S1). For models that included chronological age, sex, and GM ages as covariates, the addition of WM age nonrigid resulted in improvements in both the C-index and the AIC, indicating improved predictive accuracy and model fit, respectively (Table 3). Goodness-of-fit improvements resulted from the inclusion of WM age nonrigid were statistically significant (p-value < 0.05).

**Table 3. Added value of the WM age nonrigid in predicting MCI incidence.**

| Features | C-index (w/o WM age nonrigid) | C-index (w/ WM age nonrigid) | AIC (w/o WM age nonrigid) | AIC (w/ WM age nonrigid) | $\chi^2$ | p-value[†] |
|---|---|---|---|---|---|---|
| Basic: chronological age + sex | 0.65 (0.60, 0.70) | 0.72 (0.67, 0.78) | 1263.9 | 1236.5 | 29.42 | $\ll$0.001 |
| Basic + GM age (ours) | 0.71 (0.66, 0.76) | 0.73 (0.68, 0.78) | 1240.1 | 1236.3 | 5.79 | 0.016 |
| Basic + GM age (TSAN) | 0.71 (0.65, 0.76) | 0.73 (0.67, 0.78) | 1246.4 | 1237.8 | 10.58 | 0.001 |
| Basic + GM age (DBN) | 0.68 (0.63, 0.73) | 0.72 (0.67, 0.77) | 1253.8 | 1238.3 | 17.48 | $\ll$0.001 |

[†]Likelihood ratio tests were conducted by comparing the log likelihood of models with and without WM age nonrigid.

# Discussion

Current dMRI-based brain age estimation has significant overlap with structural MRI-based brain age estimation in terms of methodology, where two common approaches are (1) engineering features from the images and then using a regression model,(Gao et al., 2024; H. He et al., 2022; Wen et al., 2024) and (2) employing neural networks (typically convolutional neural networks) for representation learning on the images.(H. J. Cai et al., 2023; M. Chen et al., 2020; Gao et al., 2024; Wang et al., 2023) On the other hand, unique challenges exist in dMRI-based brain age estimation. One example is the intersite variability of dMRI data, for which researchers have proposed solutions such as transfer learning to improve model generalization.(C.-L. Chen et al., 2020)

Furthermore, we highlight two complementary directions that have emerged recently. The first is multimodal fusion, which focuses on integrating multiple imaging modalities into a single model.(H. J. Cai et al., 2023) This comprehensive approach leverages the synergistic information captured by different data types to produce an overall representation of brain aging. In contrast, the second direction is modality-specific modeling, which employs a separate model for each modality.(Wen et al., 2024) This approach generates modality-specific brain age estimates, capturing potentially distinct but interrelated neurobiological facets of aging. These modality-specific estimates offer a more nuanced view of how individual imaging measures relate to the overall aging process.

Our previous work investigated whether the unique microstructural features in dMRI could be used for brain age estimation.(Gao et al., 2024) The present study extends that work by including a larger dataset, which allows for a more comprehensive evaluation of the model.

In particular, data from participants who transitioned from CN to MCI enabled us to perform classification and prediction experiments, thus examining the model's clinical value. Additionally, we compared WM age models with GM age models and provided a preliminary exploration of their relative timing in the context of neurodegeneration. When implementing the GM age models, we noticed discrepancies in the image sizes. To maintain consistency with the literature, we chose to adhere to the original settings and implementations. We acknowledge that variations in image resolution can impact the results.

In this study, we continue to use FA and MD images because they are widely used for characterizing white matter microstructural changes in brain aging. We note, however, that other dMRI-derived measures also exist and could offer unique advantages for brain age estimation.(Roibu et al., 2023) Instead of training two separate models for males and females, we train a single model that takes the sex label as input. This approach allows the model to learn the shared underlying features between males and females, improving data efficiency.

By selectively focusing on microstructural information for brain age estimation, we can develop a potentially more sensitive and earlier biomarker for predicting neurodegenerative diseases. Specifically, by applying nonrigid (deformable) registration to mitigate the macrostructural information in diffusion MRI data, we have derived a distinctive microstructure-informed brain age (WM age nonrigid), which holds promise as an early indicator of mild cognitive impairment. However, we acknowledge that the nonrigid registration cannot eliminate macrostructural information and can introduce artifacts that may drive the brain age estimation model. For example, Dinsdale et al. found that models trained with nonlinearly registered T1w images were driven by areas around the ventricles, and thus likely by artifacts of registration.(Dinsdale et al., 2021) To investigate whether our WM age nonrigid model uses the intended microstructural information rather than macrostructural artifacts or residuals, we used Gradient-weighted Class Activation Mapping (Grad-CAM)(Selvaraju et al., 2017) to visualize the brain regions relevant for the brain age estimation. We found that the WM age nonrigid model relies on more areas beyond the ventricles, in contrast to the WM age affine model. Nonetheless, we emphasize that the Grad-CAM results are exploratory and should be interpreted as suggestive evidence rather than definitive proof. (Supplementary Materials) Other approaches to mitigate macrostructural information exist. For instance, using age-specific templates can help normalize registration artefacts across age bins.(Arthofer et al., 2024) However, it introduces additional challenges, such as variability between templates that may affect brain age estimation. Future studies are needed to explore these potential directions.

GM age (ours), WM age affine, and WM age nonrigid use the same 3D ResNet architecture, with nearly identical complexity (the difference being the number of input channels). The distinct behavior of these brain age estimates is driven by the type of information within the images. GM age (ours) uses skull-stripped T1w images affinely registered to the MNI152 template. The T1w images capture mainly macrostructural and intensity information. WM age affine uses skull-stripped FA and MD images affine-registered to the MNI152 template. The FA and MD images contain a blend of micro- and macrostructural information. WM age nonrigid uses skull-stripped FA and MD images nonrigid-registered to the MNI152 template. The FA and MD images contain mainly microstructural information, with macrostructural information mitigated. The difference in the information across estimation approaches leads to differences in the biomarkers' properties. In diagnostic group comparisons, WM age nonrigid appears older than GM age (ours) for CN participants who will transition to MCI, suggesting that microstructural changes detectable by FA and MD are already deviating from the normal aging trajectory, even when macrostructural changes are not yet evident in T1w images. Conversely, for AD participants, GM age (ours) appears older than WM age nonrigid, indicating the presence of significant macrostructural changes captured by T1w images.

In classifying populations as either CN or AD, classifiers using WM age affine achieved intermediate performance between those using WM age nonrigid and GM age (ours). This intermediate performance may be attributed to the macrostructural information preserved in the FA and MD images used by WM age affine. This macrostructural information enhances model performance relative to WM age nonrigid; however, due to its lower resolution (or contrast) compared to the macrostructural information in T1w images, it does not reach the performance level of GM age (ours). We note, however, that the differences in performance (Table 2) are mostly not statistically significant, likely due to the limited size of the paired testing data. The pattern of WM age affine's performance falling between WM age nonrigid and GM age (ours) is consistent in MCI prediction experiments. The AUC of classifiers using WM age affine is intermediate, or its peak occurs between the peak for GM age (ours) (at 0 years) and the peak for WM age nonrigid (at 4 years prior to MCI diagnosis), as observed with the random forest in the "global model" setup and logistic regression in the "time-specific models" setup.

We note the variability of CN in the original study papers (Beekly et al., 2007; Bennett et al., 2018; Bookheimer et al., 2019; Harms et al., 2018; Jack Jr. et al., 2008; Jefferson et al., 2016; Johnson et al., 2018; Koenig et al., 2020; Kötter et al., 2001; L Barnes et al., 2012; LaMontagne et al., 2019; Sacktor et al., 2017; Shock, 1984; Sudlow et al., 2015). CN is not a single, homogeneous state; rather, there may be states within the spectrum of CN that contribute to variations in brain aging. Examining these states is an important area.(Smith et al., 2019)

Our results suggest that WM age nonrigid is a potentially earlier biomarker for predicting MCI. It behaves differently from GM age in terms of the relative timing during the course of neurodegeneration (Fig. 4), implying that it may capture the unique microstructural information in dMRI. It provides added value to GM age models in predicting MCI incidence (Table 3). This added value is not exclusive to our GM age model but also extends to other GM age models, including TSAN and DeepBrainNet. Despite these improvements, the additional costs of implementing our approach–such as the extra time required for dMRI acquisition and preprocessing–should be considered. Nevertheless, WM age nonrigid represents a step toward improving the sensitivity of brain age estimation and can potentially benefit neurodegenerative disease prediction. To further evaluate its clinical value, a larger testing dataset consisting of longitudinal data points from participants at various stages of neurodegenerative disease development is required.

# Data and Code Availability

The datasets supporting the conclusions of this research are available, subject to certain restrictions. The datasets were used under agreement for this study and are therefore not publicly available. More information about the datasets can be found in the supplementary materials. The authors may provide data upon receiving reasonable request and with permission. The code is available at: https://github.com/MASILab/BRAID

# Author Contributions

**Chenyu Gao:** Conceptualization, Methodology, Software, Validation, Formal analysis, Investigation, Data Curation, Writing - Original Draft, Visualization **Michael E. Kim:** Software, Validation, Data Curation, Writing - Review & Editing **Karthik Ramadass:** Data Curation **Praitayini Kanakaraj:** Validation **Aravind R. Krishnan:** Writing - Review & Editing **Adam M. Saunders:** Validation, Writing - Review & Editing **Nancy R. Newlin:** Validation, Writing - Review & Editing **Ho Hin Lee:** Writing - Review & Editing **Qi Yang:** Writing - Review & Editing **Warren D. Taylor:** Writing - Review & Editing **Brian D. Boyd:** Writing - Review & Editing **Lori L. Beason-Held:** Resources, Writing - Review & Editing **Susan M. Resnick:** Resources, Writing - Review & Editing **Lisa L. Barnes:** Resources, Writing - Review & Editing **David A. Bennett:** Resources, Writing - Review & Editing **Marilyn S. Albert:** Resources **Katherine D. Van Schaik:** Writing - Review & Editing **Derek B. Archer:** Resources, Writing - Review & Editing **Timothy J. Hohman:** Resources, Writing - Review & Editing **Angela L. Jefferson:** Resources, Writing - Review & Editing **Ivana Išgum:** Supervision, Writing - Review & Editing **Daniel Moyer:** Supervision, Writing - Review & Editing **Yuankai Huo:** Supervision, Writing - Review & Editing **Kurt G. Schilling:** Resources, Writing - Review & Editing **Lianrui Zuo:** Writing - Review & Editing **Shunxing Bao:** Writing - Review & Editing **Nazirah Mohd Khairi:** Validation, Data Curation, Writing - Review & Editing **Zhiyuan Li:** Validation, Investigation, Data Curation, Writing - Review & Editing **Christos Davatzikos:** Supervision, Writing - Review & Editing **Bennett A. Landman:** Conceptualization, Methodology, Validation, Resources, Writing - Review & Editing, Supervision, Project administration, Funding acquisition.

# Acknowledgments

Data collection and sharing for this project was funded (in part) by the Alzheimer's Disease Neuroimaging Initiative (ADNI) (National Institutes of Health Grant U01 AG024904) and DOD ADNI (Department of Defense award number W81XWH-12-2-0012). ADNI is funded by the National Institute on Aging, the National Institute of Biomedical Imaging and Bioengineering, and through generous contributions from the following: AbbVie, Alzheimer's Association; Alzheimer's Drug Discovery Foundation; Araclon Biotech; BioClinica, Inc.; Biogen; Bristol-Myers Squibb Company; CereSpir, Inc.; Cogstate; Eisai Inc.; Elan Pharmaceuticals, Inc.; Eli Lilly and Company; EuroImmun; F. Hoffmann-La Roche Ltd and its affiliated company Genentech, Inc.; Fujirebio; GE Healthcare; IXICO Ltd.; Janssen Alzheimer Immunotherapy Research & Development, LLC.; Johnson & Johnson Pharmaceutical Research & Development LLC.; Lumosity; Lundbeck; Merck & Co., Inc.; Meso Scale Diagnostics, LLC.; NeuroRx Research; Neurotrack Technologies; Novartis Pharmaceuticals Corporation; Pfizer Inc.; Piramal Imaging; Servier; Takeda Pharmaceutical Company; and Transition Therapeutics. The Canadian Institutes of Health Research is providing funds to support ADNI clinical sites in Canada. Private sector contributions are facilitated by the Foundation for the National Institutes of Health (www.fnih.org). The grantee organization is the Northern California Institute for Research and Education, and the study is coordinated by the Alzheimer's Therapeutic Research Institute at the University of Southern California. ADNI data are disseminated by the Laboratory for Neuro Imaging at the University of Southern California.

The BIOCARD study is supported by a grant from the National Institute on Aging (NIA): U19-AG03365. The BIOCARD Study consists of 7 Cores and 2 projects with the following members: (1) The Administrative Core (Marilyn Albert, Corinne Pettigrew, Barbara Rodzon);

(2) the Clinical Core (Marilyn Albert, Anja Soldan, Rebecca Gottesman, Corinne Pettigrew, Leonie Farrington, Maura Grega, Gay Rudow, Rostislav Brichko, Scott Rudow, Jules Giles, Ned Sacktor); (3) the Imaging Core (Michael Miller, Susumu Mori, Anthony Kolasny, Hanzhang Lu, Kenichi Oishi, Tilak Ratnanather, Peter vanZijl, Laurent Younes); (4) the Biospecimen Core (Abhay Moghekar, Jacqueline Darrow, Alexandria Lewis, Richard O'Brien); (5) the Informatics Core (Roberta Scherer, Ann Ervin, David Shade, Jennifer Jones, Hamadou Coulibaly, Kathy Moser, Courtney Potter); the (6) Biostatistics Core (Mei-Cheng Wang, Yuxin Zhu, Jiangxia Wang); (7) the Neuropathology Core (Juan Troncoso, David Nauen, Olga Pletnikova, Karen Fisher); (8) Project 1 (Paul Worley, Jeremy Walston, Mei-Fang Xiao), and (9) Project 2 (Mei-Cheng Wang, Yifei Sun, Yanxun Xu.

The BLSA is supported by the Intramural Research Program, National Institute on Aging, NIH.

Data collection and sharing for this project was provided by the Human Connectome Project (HCP; PI: Bruce Rosen, M.D., Ph.D., Arthur W. Toga, Ph.D., Van J. Weeden, MD). HCP funding was provided by the National Institute of Dental and Craniofacial Research (NIDCR), the National Institute of Mental Health (NIMH), and the National Institute of Neurological Disorders and Stroke (NINDS). HCP data are disseminated by the Laboratory of Neuro Imaging at the University of Southern California.

Data collection and sharing for this project was provided by the International Consortium for Brain Mapping (ICBM; Principal Investigator: John Mazziotta, MD, PhD). ICBM funding was provided by the National Institute of Biomedical Imaging and BioEngineering. ICBM data are disseminated by the Laboratory of Neuro Imaging at the University of Southern California.

The NACC database is funded by NIA/NIH Grant U24 AG072122. NACC data are contributed by the NIA-funded ADRCs: P30 AG062429 (PI James Brewer, MD, PhD), P30 AG066468 (PI Oscar Lopez, MD), P30 AG062421 (PI Bradley Hyman, MD, PhD), P30 AG066509 (PI Thomas Grabowski, MD), P30 AG066514 (PI Mary Sano, PhD), P30 AG066530 (PI Helena Chui, MD), P30 AG066507 (PI Marilyn Albert, PhD), P30 AG066444 (PI John Morris, MD), P30 AG066518 (PI Jeffrey Kaye, MD), P30 AG066512 (PI Thomas Wisniewski, MD), P30 AG066462 (PI Scott Small, MD), P30 AG072979 (PI David Wolk, MD), P30 AG072972 (PI Charles DeCarli, MD), P30 AG072976 (PI Andrew Saykin, PsyD), P30 AG072975 (PI Julie Schneider, MD), P30 AG072978 (PI Neil Kowall, MD), P30 AG072977 (PI Robert Vassar, PhD), P30 AG066519 (PI Frank LaFerla, PhD), P30 AG062677 (PI Ronald Petersen, MD, PhD), P30 AG079280 (PI Eric Reiman, MD), P30 AG062422 (PI Gil Rabinovici, MD), P30 AG066511 (PI Allan Levey, MD, PhD), P30 AG072946 (PI Linda Van Eldik, PhD), P30 AG062715 (PI Sanjay Asthana, MD, FRCP), P30 AG072973 (PI Russell Swerdlow, MD), P30 AG066506 (PI Todd Golde, MD, PhD), P30 AG066508 (PI Stephen Strittmatter, MD, PhD), P30 AG066515 (PI Victor Henderson, MD, MS), P30 AG072947 (PI Suzanne Craft, PhD), P30 AG072931 (PI Henry Paulson, MD, PhD), P30 AG066546 (PI Sudha Seshadri, MD), P20 AG068024 (PI Erik Roberson, MD, PhD), P20 AG068053 (PI Justin Miller, PhD), P20 AG068077 (PI Gary Rosenberg, MD), P20 AG068082 (PI Angela Jefferson, PhD), P30 AG072958 (PI Heather Whitson, MD), P30 AG072959 (PI James Leverenz, MD).

Data were provided by OASIS-3 and OASIS-4. OASIS-3: Longitudinal Multimodal Neuroimaging: Principal Investigators: T. Benzinger, D. Marcus, J. Morris; NIH P30 AG066444, P50 AG00561, P30 NS09857781, P01 AG026276, P01 AG003991, R01

AG043434, UL1 TR000448, R01 EB009352. AV-45 doses were provided by Avid Radiopharmaceuticals, a wholly owned subsidiary of Eli Lilly. OASIS-4: Clinical Cohort: Principal Investigators: T. Benzinger, L. Koenig, P. LaMontagne.

Data contributed from ROS/MAP/MARS was supported by NIA R01AG017917, P30AG10161, P30AG072975, R01AG022018, R01AG056405, UH2NS100599, UH3NS100599, R01AG064233, R01AG15819 and R01AG067482, and the Illinois Department of Public Health (Alzheimer's Disease Research Fund). Data can be accessed at www.radc.rush.edu.

Study data were obtained from the Vanderbilt Memory and Aging Project (VMAP). VMAP data were collected by Vanderbilt Memory and Alzheimer's Center Investigators at Vanderbilt University Medical Center. This work was supported by NIA grants R01-AG034962 (PI: Jefferson), R01-AG056534 (PI: Jefferson), R01-AG062826 (PI: Gifford), U19-AG03655 (PI:Albert) and Alzheimer's Association IIRG-08-88733 (PI: Jefferson).

The data contributed from the Wisconsin Registry for Alzheimer's Prevention was supported by NIA AG021155, AG0271761, AG037639, and AG054047.

# Funding

This project is funded by NIH grant 1R01EB017230, U24AG074855, R01MH121620, K01EB032898, K01-AG073584. The Vanderbilt Institute for Clinical and Translational Research (VICTR) is funded by the National Center for Advancing Translational Sciences (NCATS) Clinical Translational Science Award (CTSA) Program, Award Number 5UL1TR002243-03. The content is solely the responsibility of the authors and does not necessarily represent the official views of the NIH. This work was conducted in part using the resources of the Advanced Computing Center for Research and Education at Vanderbilt University, Nashville, TN.

# Declaration of Competing Interests

The authors (Chenyu Gao, Bennett A. Landman, and Michael E. Kim) declare that they have filed a provisional patent application (U.S. Patent 63/701,861, System and Method of Brain Age Identification for Predicting Neuro-Degenerative Disease, filed October 1, 2024) related to the methods described in this manuscript.

# Supplementary materials

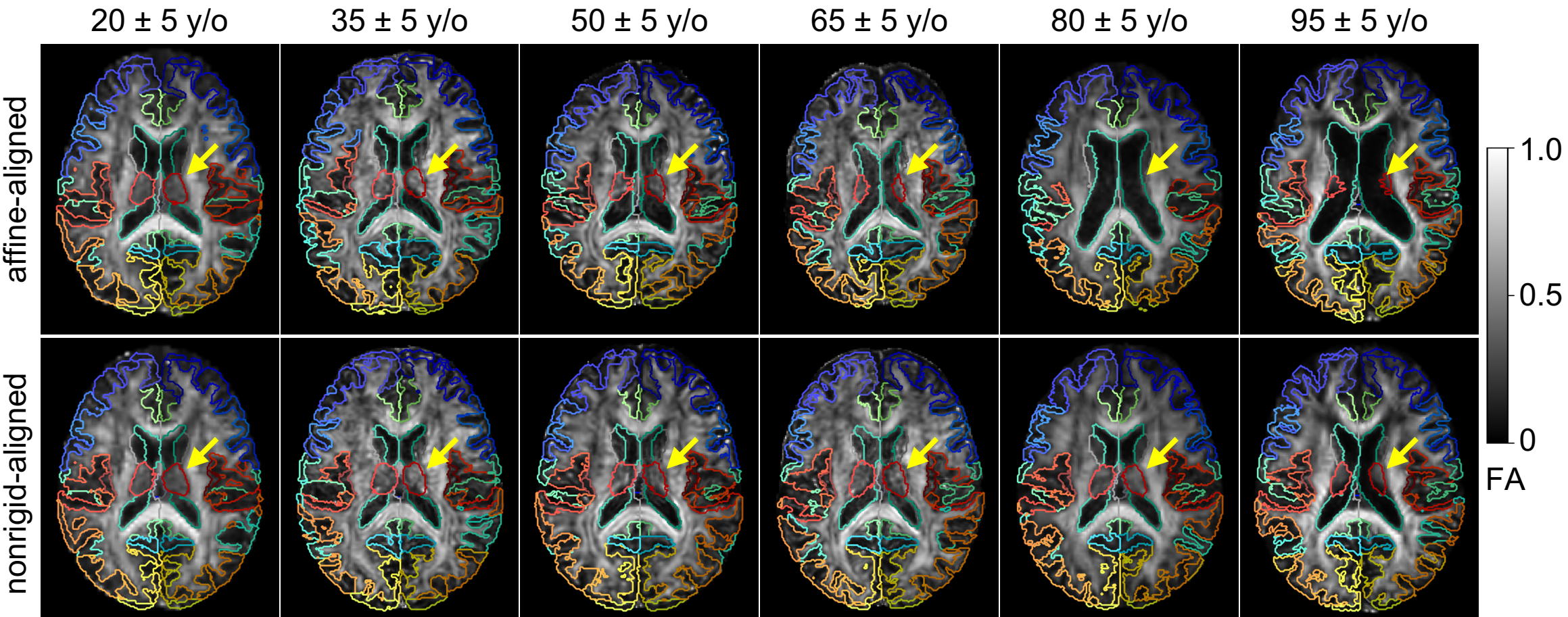

**Figure S1.** The macrostructural variations are present in the affine-aligned fractional anisotropy (FA) images, while mitigated in the nonrigid-aligned images. Contours of regions are provided to assist in the visual inspection of brain region shapes. Yellow arrows indicate the thalamus, which appears to shrink with age in the first row (affine-aligned) but remains consistent in shape and size in the second row (nonrigid-aligned).

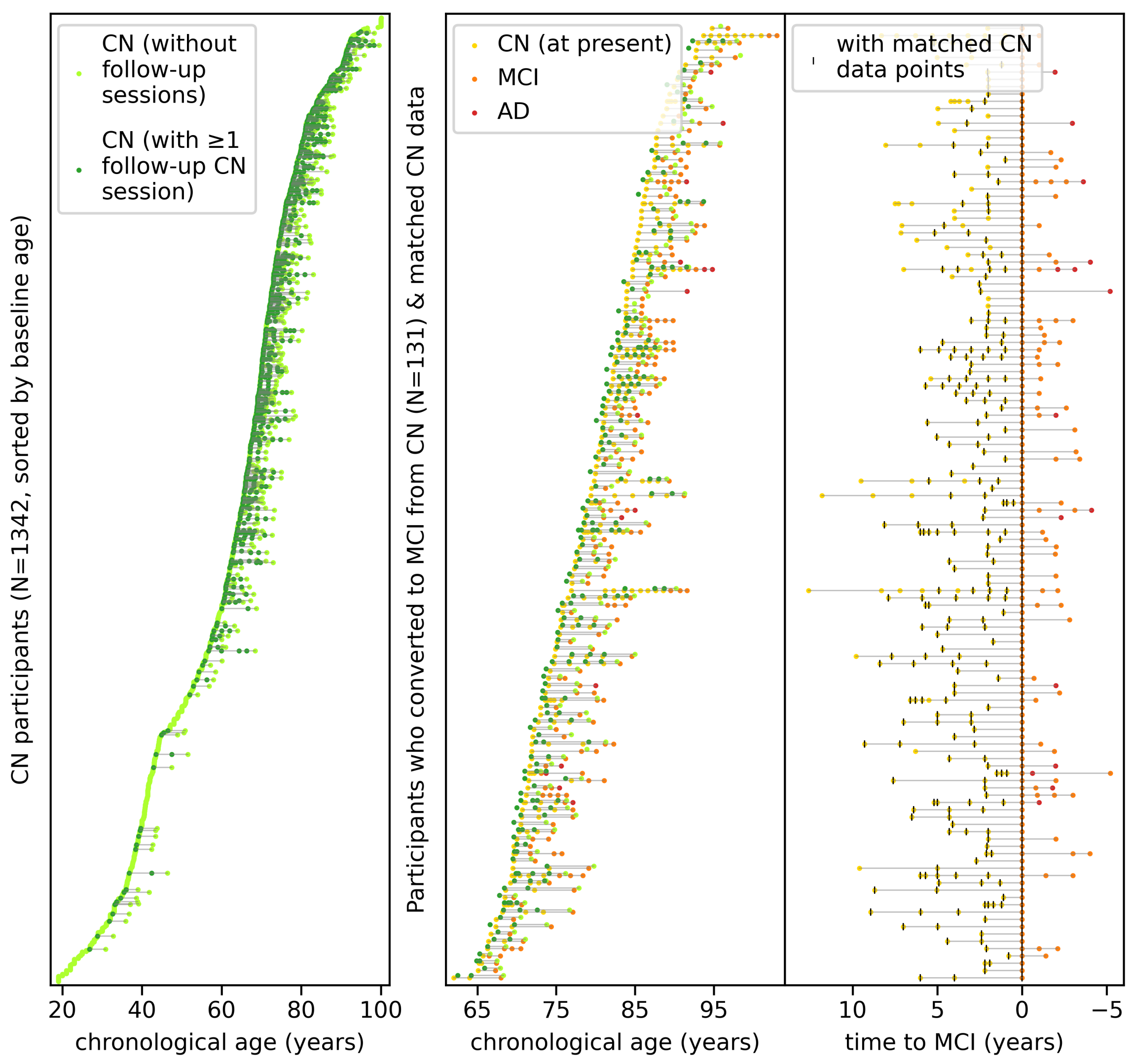

**Figure S2.** We match CN data points for participants who converted from CN to MCI based on sex, age, and time to event (i.e., time to first MCI diagnosis for MCI participants and time to last CN session for CN participants).

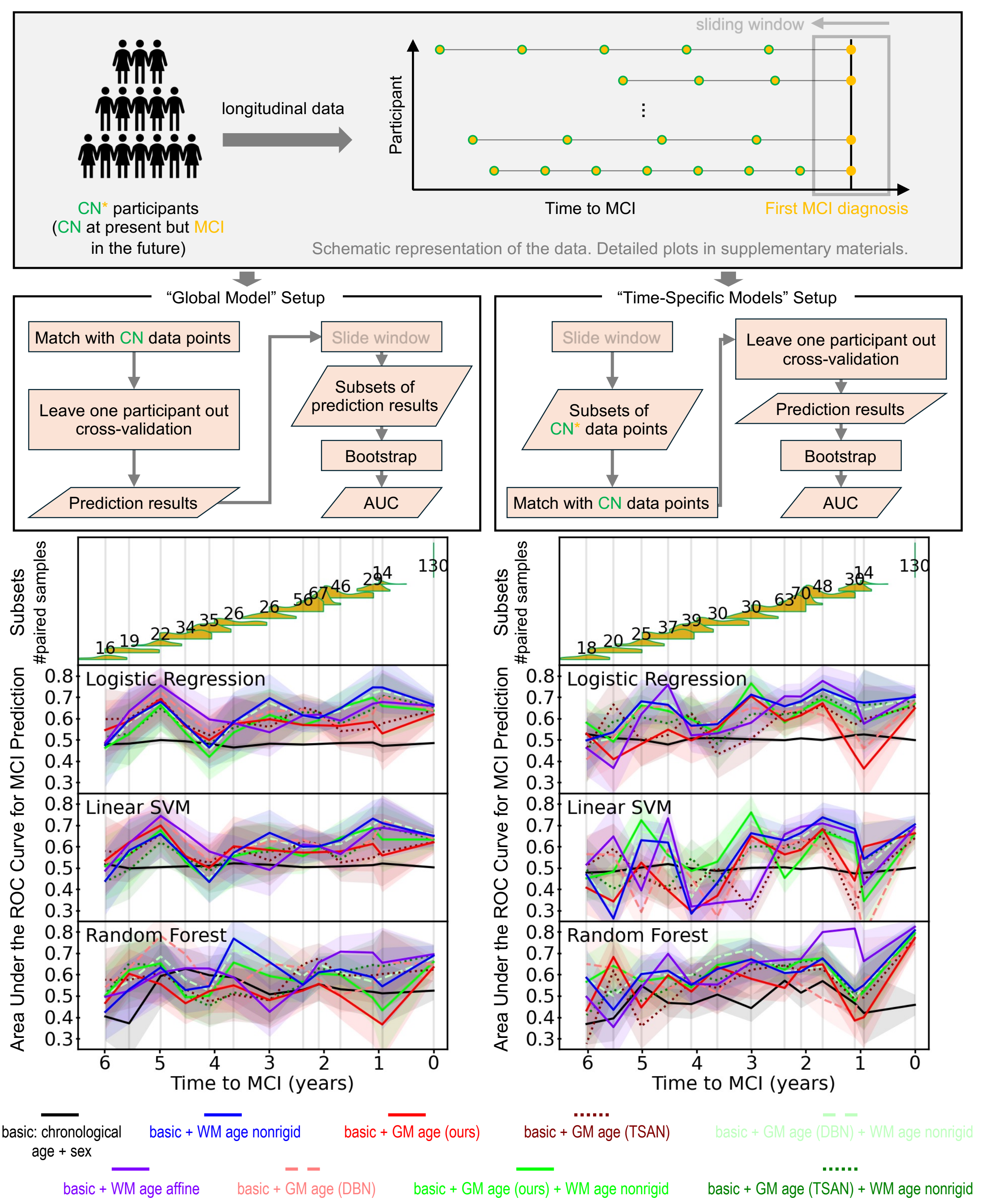

**Figure S3.** The longitudinal data from CN* participants are used for MCI prediction from *n* years pre-diagnosis in two experimental setups. In the “Global Model” setup, WM age nonrigid shows an advantage from 0 to approximately 3.5 years before MCI. In the “Time-Specific Models” setup, WM age nonrigid shows an advantage up to approximately 4–5 years before MCI. However, these advantages are not statistically significant, as indicated by the overlapping confidence intervals.

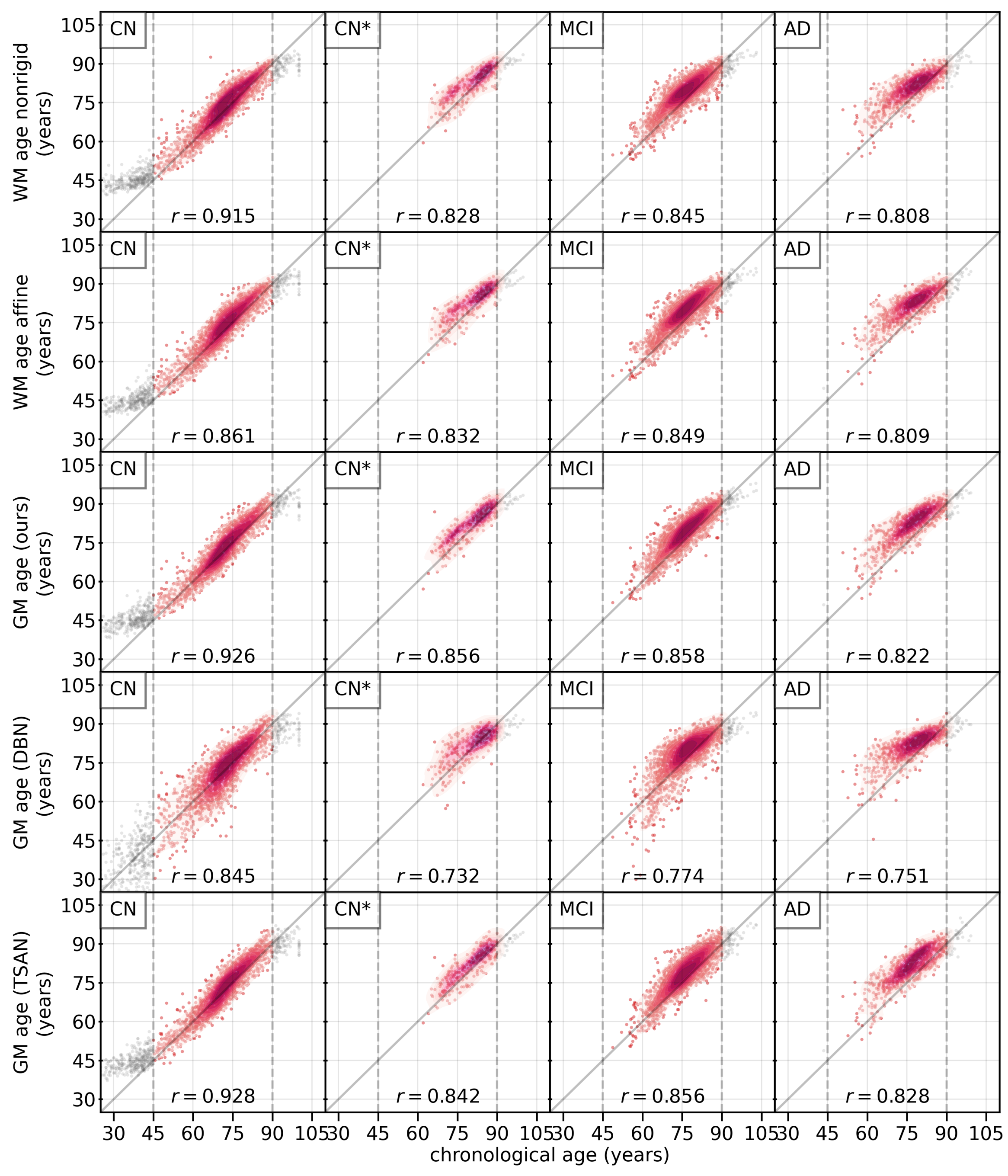

**Figure S4.** Estimated brain age vs. chronological age for participants that are (from left to right column) CN (cognitively normal), CN* (cognitively normal at present but transitioning to mild cognitive impairment), MCI (mild cognitive impairment), and AD (Alzheimer's disease). The Pearson correlation coefficients are calculated on data points within the age range of 45 to 90 years.

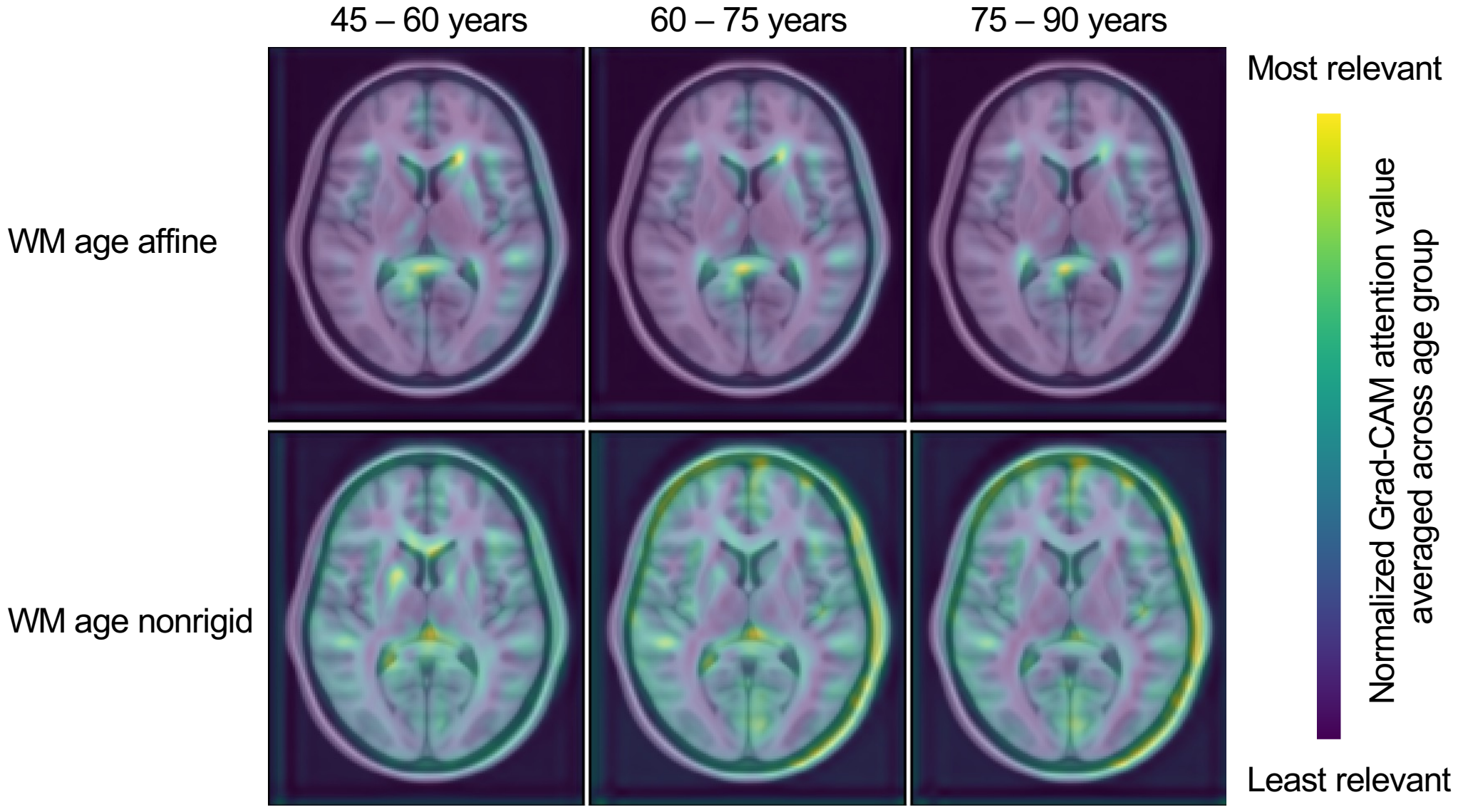

**Figure S5.** We compute Grad-CAM attention for WM age affine and WM age nonrigid, respectively. The attention values are normalized for each image, averaged across images from the same age group, and overlaid on the MNI152 template image. WM age nonrigid is driven by more regions beyond the ventricles, in contrast to WM age affine.

**Table S1.** Life table for the survival analysis.

| Interval (years) | n CN at Beginning of Interval | n MCI During Interval | n Censored |
|---|---|---|---|
| 0-2 | 421 | 19 | 29 |
| 2-4 | 373 | 49 | 127 |
| 4-6 | 197 | 34 | 86 |
| 6-8 | 77 | 17 | 26 |
| 8-10 | 34 | 10 | 21 |
| 10-12 | 3 | 1 | 1 |
| 12-14 | 1 | 1 | 0 |

**Table S2.** Classification of CN vs. AD, CN vs. MCI, and CN vs. CN* using chronological age, sex, and brain age-related features. To facilitate easier visual inspection, we use blue for WM age nonrigid, purple for WM age affine, red for GM ages, and green for combinations of GM ages and WM age nonrigid. The same color scheme is followed in other figures.

| | CN vs. AD (N=458 matched pairs) | | | | | |
|---|---|---|---|---|---|---|
| | Logistic Regression | | Linear SVM | | Random Forest | |
| Features | Accuracy | AUC | Accuracy | AUC | Accuracy | AUC |
| basic: chronological age + sex | 0.50 (0.50, 0.50) | 0.50 (0.50, 0.50) | 0.50 (0.50, 0.50) | 0.50 (0.50, 0.50) | 0.56 (0.55, 0.58) | 0.59 (0.57, 0.60) |
| + WM age nonrigid | 0.65 (0.62, 0.67) | 0.72 (0.70, 0.74) | 0.65 (0.63, 0.67) | 0.72 (0.70, 0.74) | 0.69 (0.67, 0.72) | 0.77 (0.74, 0.79) |
| + WM age affine | 0.67 (0.64, 0.69) | 0.74 (0.72, 0.77) | 0.67 (0.65, 0.70) | 0.74 (0.72, 0.76) | 0.69 (0.67, 0.71) | 0.77 (0.74, 0.79) |
| + GM age (ours) | 0.69 (0.67, 0.71) | 0.76 (0.74, 0.79) | 0.69 (0.66, 0.71) | 0.76 (0.73, 0.78) | 0.70 (0.68, 0.73) | 0.78 (0.76, 0.80) |
| + GM age (DBN) | 0.70 (0.68, 0.73) | 0.78 (0.76, 0.80) | 0.70 (0.68, 0.73) | 0.78 (0.76, 0.80) | 0.69 (0.67, 0.71) | 0.77 (0.75, 0.79) |
| + GM age (TSAN) | 0.68 (0.66, 0.71) | 0.76 (0.74, 0.78) | 0.68 (0.66, 0.71) | 0.75 (0.73, 0.78) | 0.68 (0.66, 0.70) | 0.76 (0.74, 0.78) |
| + GM age (ours) + WM age nonrigid | 0.70 (0.68, 0.72) | 0.76 (0.74, 0.78) | 0.70 (0.67, 0.72) | 0.75 (0.73, 0.78) | 0.73 (0.71, 0.75) | 0.79 (0.77, 0.82) |
| + GM age (DBN) + WM age nonrigid | **0.70 (0.68, 0.73)** | **0.79 (0.77, 0.81)** | **0.71 (0.69, 0.74)** | **0.79 (0.76, 0.81)** | **0.74 (0.72, 0.76)** | **0.81 (0.79, 0.83)** |
| + GM age (TSAN) + WM age nonrigid | 0.68 (0.66, 0.70) | 0.76 (0.74, 0.78) | 0.70 (0.68, 0.72) | 0.76 (0.73, 0.78) | 0.72 (0.69, 0.74) | 0.79 (0.77, 0.81) |

| | CN vs. MCI (N=694 matched pairs) | | | | | |
|---|---|---|---|---|---|---|
| | Logistic Regression | | Linear SVM | | Random Forest | |
| Features | Accuracy | AUC | Accuracy | AUC | Accuracy | AUC |
| basic: chronological age + sex | 0.50 (0.50, 0.50) | 0.50 (0.50, 0.50) | 0.50 (0.50, 0.50) | 0.50 (0.50, 0.50) | 0.57 (0.56, 0.59) | 0.59 (0.58, 0.61) |
| + WM age nonrigid | 0.61 (0.59, 0.63) | 0.65 (0.62, 0.67) | 0.60 (0.59, 0.62) | 0.65 (0.62, 0.67) | 0.64 (0.62, 0.66) | 0.70 (0.68, 0.72) |
| + WM age affine | 0.59 (0.57, 0.61) | 0.63 (0.61, 0.65) | 0.60 (0.58, 0.62) | 0.63 (0.61, 0.65) | 0.66 (0.64, 0.68) | 0.69 (0.67, 0.71) |
| + GM age (ours) | 0.62 (0.60, 0.63) | 0.65 (0.63, 0.67) | 0.60 (0.58, 0.62) | 0.64 (0.61, 0.66) | 0.67 (0.65, 0.69) | 0.72 (0.70, 0.74) |
| + GM age (DBN) | 0.59 (0.57, 0.61) | 0.62 (0.59, 0.64) | 0.58 (0.56, 0.60) | 0.61 (0.59, 0.63) | 0.59 (0.57, 0.61) | 0.63 (0.61, 0.65) |
| + GM age (TSAN) | 0.59 (0.57, 0.61) | 0.63 (0.61, 0.65) | 0.57 (0.56, 0.60) | 0.62 (0.60, 0.64) | 0.63 (0.61, 0.65) | 0.68 (0.65, 0.70) |
| + GM age (ours) + WM age nonrigid | **0.62 (0.60, 0.64)** | **0.66 (0.64, 0.68)** | **0.63 (0.61, 0.65)** | **0.66 (0.64, 0.68)** | **0.67 (0.65, 0.69)** | **0.73 (0.71, 0.75)** |
| + GM age (DBN) + WM age nonrigid | 0.61 (0.59, 0.63) | 0.65 (0.63, 0.67) | 0.61 (0.59, 0.62) | 0.65 (0.63, 0.67) | 0.66 (0.64, 0.68) | 0.72 (0.70, 0.74) |
| + GM age (TSAN) + WM age nonrigid | 0.60 (0.58, 0.62) | 0.65 (0.63, 0.67) | 0.60 (0.58, 0.62) | 0.64 (0.62, 0.67) | 0.66 (0.65, 0.68) | 0.71 (0.69, 0.73) |

| | CN vs. CN* (N=118 matched pairs) | | | | | |
|---|---|---|---|---|---|---|
| | Logistic Regression | | Linear SVM | | Random Forest | |
| Features | Accuracy | AUC | Accuracy | AUC | Accuracy | AUC |
| basic: chronological age + sex | 0.50 (0.50, 0.51) | 0.50 (0.50, 0.50) | 0.50 (0.50, 0.50) | 0.50 (0.50, 0.50) | 0.53 (0.50, 0.57) | 0.56 (0.52, 0.60) |
| + WM age nonrigid | 0.60 (0.56, 0.65) | 0.64 (0.59, 0.69) | **0.62 (0.57, 0.67)** | 0.64 (0.59, 0.69) | 0.61 (0.56, 0.66) | 0.63 (0.58, 0.68) |
| + WM age affine | 0.61 (0.56, 0.66) | **0.64 (0.59, 0.70)** | 0.59 (0.54, 0.64) | 0.63 (0.58, 0.69) | 0.55 (0.50, 0.59) | 0.60 (0.55, 0.66) |
| + GM age (ours) | 0.60 (0.56, 0.65) | 0.64 (0.59, 0.69) | 0.62 (0.57, 0.67) | 0.63 (0.57, 0.69) | 0.60 (0.55, 0.64) | 0.63 (0.57, 0.68) |
| + GM age (DBN) | 0.56 (0.51, 0.61) | 0.62 (0.56, 0.67) | 0.57 (0.52, 0.62) | 0.61 (0.56, 0.66) | 0.57 (0.53, 0.62) | 0.59 (0.54, 0.65) |
| + GM age (TSAN) | 0.57 (0.53, 0.62) | 0.62 (0.57, 0.68) | 0.58 (0.54, 0.63) | 0.61 (0.56, 0.67) | 0.57 (0.52, 0.62) | 0.58 (0.52, 0.63) |
| + GM age (ours) + WM age nonrigid | 0.60 (0.55, 0.65) | 0.64 (0.59, 0.69) | 0.61 (0.56, 0.65) | 0.64 (0.59, 0.69) | 0.61 (0.56, 0.65) | 0.65 (0.60, 0.70) |
| + GM age (DBN) + WM age nonrigid | **0.62 (0.57, 0.66)** | 0.64 (0.59, 0.69) | 0.60 (0.56, 0.65) | **0.65 (0.59, 0.70)** | **0.61 (0.57, 0.66)** | **0.66 (0.61, 0.71)** |
| + GM age (TSAN) + WM age nonrigid | 0.60 (0.55, 0.64) | 0.64 (0.59, 0.69) | 0.58 (0.53, 0.63) | 0.63 (0.58, 0.68) | 0.61 (0.56, 0.65) | 0.65 (0.59, 0.69) |

CN = cognitively normal; AD = Alzheimer's disease; MCI = mild cognitive impairment; CN* = cognitively normal at present but diagnosed with mild cognitive impairment in the future.

# Dataset descriptions

- **ADNI**: Data used in the preparation of this article were obtained from the Alzheimer's Disease Neuroimaging Initiative (ADNI) database (adni.loni.usc.edu). The ADNI was launched in 2003 as a public-private partnership, led by Principal Investigator Michael W. Weiner, MD. The primary goal of ADNI has been to test whether serial magnetic resonance imaging (MRI), positron emission tomography (PET), other biological markers, and clinical and neuropsychological assessment can be combined to measure the progression of mild cognitive impairment (MCI) and early Alzheimer's disease (AD). Information about the dMRI sequence used in the present study is provided below. For further details, please visit https://adni.loni.usc.edu/

| TE (s) | TR (s) | Number of directions (b=0) | Shells | Manufacturer | Model type | Field Strength |
|---|---|---|---|---|---|---|
| 0.08 | 14.2 | 46 (5) | 0, 1000 | GE | Signa HDxt | 3T |
| 0.06 | 9.1 | 46 (5) | 0, 1000 | GE | Discovery MR750 | 3T |
| 0.06 | 7.8 | 54 (6) | 0, 1000 | GE | Discovery MR750 | 3T |
| 0.06 | 7.8 | 54 (6) | 0, 1000 | GE | Discovery MR750 | 3T |
| 0.09 | 10.1 | 33 (1) | 0, 1000 | Philips | Achieva dStream | 3T |
| 0.06 | 9.0 | 36 (4) | 0, 1000 | GE | Signa Premier | 3T |
| 0.08 | 9.6 | 55 (7) | 0, 1000 | Siemens | Skyra | 3T |
| 0.06 | 7.2 | 55 (6) | 0, 1000 | Siemens | Prisma_fit | 3T |
| 0.06 | 7.8 | 54 (6) | 0, 1000 | GE | DISCOVERY MR750 | 3T |
| 0.06 | 7.8 | 54 (6) | 0, 1000 | GE | DISCOVERY MR750 | 3T |
| 0.08 | 15.3 | 36 (4) | 0, 1000 | GE | DISCOVERY MR750w | 3T |
| 0.07 | 13.0 | 46 (5) | 0, 1000 | GE | Signa HDxt | 3T |
| 0.07 | 3.4 | 127 (13) | 0, 500, 1000, 2000 | Siemens | Prisma | 3T |
| 0.06 | 7.2 | 55 (6) | 0, 1000 | Siemens | Prisma_fit | 3T |
| 0.07 | 3.4 | 127 (13) | 0, 500, 1000, 2000 | Siemens | Prisma | 3T |
| 0.10 | 10.9 | 36 (4) | 0, 1000 | Philips | Ingenia | 3T |
| 0.07 | 3.4 | 127 (12) | 0, 500, 1000, 2000 | Siemens | Prisma | 3T |
| 0.07 | 12.5 | 46 (5) | 0, 1000 | GE | Signa HDxt | 3T |
| 0.06 | 7.2 | 55 (7) | 0, 1000 | Siemens | Prisma_fit | 3T |
| 0.07 | 9.1 | 46 (5) | 0, 1000 | GE | DISCOVERY MR750 | 3T |
| 0.06 | 9.1 | 46 (5) | 0, 1000 | GE | DISCOVERY MR750 | 3T |

- **BIOCARD**: Information about the dMRI sequence used in the present study is provided below. For further details, please visit https://www.biocard-se.org/

| TE (s) | TR (s) | Number of directions (b=0) | Shells | Manufacturer | Model type | Field Strength |
|---|---|---|---|---|---|---|
| 0.075 | 7.5 | 33 (1) | 0, 700 | Phillips | Achieva | 3T |

- **BLSA**: Information about the dMRI sequence used in the present study is provided below. For further details, please visit https://blsa.nih.gov/

| | Scanner A | Scanners B/C | Scanner D |
|---|---|---|---|
| Head coil | Philips 8-ch | Philips 8-ch | Philips 8-ch |
| Scan time (mins:secs) | 3:56 | 3:58 | 4:20 |
| Number of gradients | 30 | 32 | 32 |
| Number of b0 images | 1 | 1 | 1 |
| Max b-factor (s/mm²) | 700 | 700 | 700 |
| Number of signal averages (NSA) | 1 | 1 | 1 |
| Diffusion gradient timing DELTA/delta (ms) | 39.2/15.1 | 36.3/16 | 36.3/13.5 |
| Slice thickness (mm) | 2.5 | 2.2 | 2.2 |
| Number of slices | 50 | 65 | 70 |
| Flip angle (deg) | 90 | 90 | 90 |
| TR/TE (ms) | 6210/80 | 6801/75 | 7454/75 |
| Field of view (mm) | 240x240 | 212x212 | 260x260 |
| Acquisition matrix | 96x96 | 96x95 | 116x115 |
| Reconstruction matrix | 256x256 | 256x256 | 320x320 |
| Reconstructed voxel size (mm) | 0.94x0.94 | 0.83x0.83 | 0.81x0.81 |

- **HCPA**: Data used in the preparation of this work were obtained from the Human Connectome Project (HCP) database (https://ida.loni.usc.edu/login.jsp). The HCP project (Principal Investigators: Bruce Rosen, M.D., Ph.D., Martinos Center at Massachusetts General Hospital; Arthur W. Toga, Ph.D., University of Southern California, Van J. Weeden, MD, Martinos Center at Massachusetts General Hospital) is supported by the National Institute of Dental and Craniofacial Research (NIDCR), the National Institute of Mental Health (NIMH) and the National Institute of Neurological Disorders and Stroke (NINDS). HCP is the result of efforts of co-investigators from the University of Southern California, Martinos Center for Biomedical Imaging at Massachusetts General Hospital (MGH), Washington University, and the University of Minnesota. (https://www.humanconnectome.org/study/hcp-lifespan-aging) Information about the imaging protocols can be found at https://www.humanconnectome.org/study/hcp-lifespan-aging/project-protocol/imaging-protocols-hcp-aging
- **ICBM**: Data used in the preparation of this work were obtained from the International Consortium for Brain Mapping (ICBM) database (www.loni.usc.edu/ICBM). The ICBM project (Principal Investigator John Mazziotta, M.D., University of California, Los Angeles) is supported by the National Institute of Biomedical Imaging and BioEngineering. ICBM is the result of efforts of co-investigators from UCLA, Montreal Neurologic Institute, University of Texas at San Antonio, and the Institute of Medicine, Juelich/Heinrich Heine University - Germany. (www.loni.usc.edu/ICBM)

- **NACC**: Information about the dMRI sequence used in the present study is provided below. For further details, please visit https://www.naccdata.org/

| TE (s) | TR (s) | Number of directions (b=0) | Shells | Manufacturer | Model type | Field Strength |
|---|---|---|---|---|---|---|
| 0.06 | 9.1 | 7 (1) | 0, 1000 | GE | DISCOVERY_MR750 | 3T |
| 0.06 | 8.0 | 26 (1) | 0, 1000 | GE | DISCOVERY_MR750 | 3T |
| 0.10 | 9.5 | 92 (80) | 0, 1000 | Siemens | TrioTim | 3T |
| 0.08 | 8.0 | 48 (8) | 0, 1300 | GE | DISCOVERY_MR750 | 3T |
| 0.09 | 8.0 | 26 (2) | 0, 1000 | GE | GENESIS_SIGNA | 1.5T |
| 0.10 | 8.8 | 65 (1) | 0, 3000 | Phillips | Achieva | 3T |

- **OASIS3** and **OASIS4**: Information about the dMRI sequence used in the present study is provided below. For further details, please visit https://sites.wustl.edu/oasisbrains/

| TE (s) | TR (s) | Number of directions (b=0) | Shells | Manufacturer | Model type | Field Strength |
|---|---|---|---|---|---|---|
| 0.10 | 3.8 | 21 (3) | 0, 1000 | Siemens | Skyra | 3T |

- **ROSMAPMARS**: Information about the dMRI sequence used in the present study is provided below. For further details, please visit https://www.rushu.rush.edu/research/departmental-research/rush-alzheimers-disease-center/rush-alzheimers-disease-center-research/epidemologic-research

| TE (s) | TR (s) | Number of directions (b=0) | Shells | Manufacturer | Model type | Field Strength |
|---|---|---|---|---|---|---|
| 0.09 | 5.4 | 84 (12) | 0, 900 | GE | SIGNA_EXCITE | 1.5T |
| 0.09 | 8.1 | 46 (6) | 0, 1000 | Siemens | TrioTim | 3T |
| 0.09 | 8.1 | 46 (6) | 0, 1000 | Siemens | TrioTim | 3T |
| 0.09 | 8.1 | 46 (6) | 0, 1000 | Siemens | TrioTim | 3T |
| 0.05 | 12.0 | 41 (1) | 0, 1000 | PHILIPS-F398EB2 | Achieva | 3T |
| 0.05 | 11.0 | 41 (1) | 0, 1000 | PHILIPS-IJI1EMU | Achieva_dStream | 3T |
| 0.05 | 11.5 | 41 (1) | 0, 1000 | PHILIPS-EKUR50U | Achieva_dStream | 3T |

- **UKBB**: The UK Biobank subset used in the present study includes all diffusion MRI and T1-weighted MRI scans approved and provided to us through application ID 16315 (Project title: Population Mapping of Brain, Eye, Spinal Cord, and Abdomen Anatomy in the Context of Electronic Medical Records), approved on April 6th, 2018. Information about the dMRI sequence used in the present study is provided below. For further details, please visit https://www.ukbiobank.ac.uk/

Resolution: 2x2x2 mm
Field-of-view: 104x104x72 matrix
Duration: 7 minutes (including 36 seconds phase-encoding reversed data)
5x b=0 (+3x b=0 blip-reversed), 50x b=1000 s/mm2, 50x b=2000 s/mm2
Gradient timings: δ=21.4 ms, Δ=45.5 ms; Spoiler b-value = 3.3 s/mm2 SE-EPI with x3 multislice acceleration, no iPAT, fat saturation

For the two diffusion-weighted shells, 50 distinct diffusion-encoding directions were acquired (and all 100 directions are distinct). The diffusion prepraration is a standard ("monopolar") Stejskal-Tanner pulse sequence. This enables higher SNR due to a shorter echo time (TE=92ms) than than a twice-refocused ("bipolar") sequence. This improvement comes at the expense of stronger eddy current distortions, which are removed in the image processing pipeline.

(https://biobank.ctsu.ox.ac.uk/crystal/crystal/docs/brain_mri.pdf)

- **VMAP**: Information about the dMRI sequence used in the present study is provided below. For further details, please visit https://www.vumc.org/vmac/vmap

| TE (s) | TR (s) | Number of directions (b=0) | Shells | Manufacturer | Model type | Field Strength |
|---|---|---|---|---|---|---|
| 0.06 | 10 | 37 (5) | 0,1000 | Philips | Achieva | 3T |

- **WRAP**: Information about the dMRI sequence used in the present study is provided below. For further details, please visit https://wrap.wisc.edu/

| TE (s) | TR (s) | Number of directions (b=0) | Shells | Manufacturer | Model type | Field Strength |
|---|---|---|---|---|---|---|
| 0.07 | 8.0 | 48 (8) | 0, 1300 | GE | DISCOVERY MR750 | 3T |